# AI and Its Impact on Creativity and Diversity: An Empirical Study of LLM-Generated Product Ideas

Christian Terwiesch (corresponding author)

Operations, Information and Decisions Department, The Wharton School, University of Pennsylvania
terwiesch@wharton.upenn.edu
500 Jon M. Huntsman Hall, 3730 Walnut Street, Philadelphia, PA 19104

Lennart Meincke

WHU – Otto Beisheim School of Management

Mack Institute for Innovation Management, The Wharton School, University of Pennsylvania
lennart@wharton.upenn.edu
3620 Locust Walk Ste 3400, Philadelphia, PA 19104

Karan Girotra

Cornell Tech and the Cornell SC Johnson College of Business, Cornell University
kg488@cornell.edu
2 West Loop Road, New York, NY 10044

Ethan Mollick

Management Department, The Wharton School, University of Pennsylvania
emollick@wharton.upenn.edu
2000 Steinberg Hall-Dietrich Hall, 3620 Locust Walk, Philadelphia, PA 19104

Gideon Nave

Marketing Department, The Wharton School, University of Pennsylvania
gnave@wharton.upenn.edu
700 Jon M. Huntsman Hall, 3730 Walnut Street, Philadelphia, PA 19104

Karl Ulrich

Operations, Information and Decisions Department, The Wharton School, University of Pennsylvania
terwiesch@wharton.upenn.edu
500 Jon M. Huntsman Hall, 3730 Walnut Street, Philadelphia, PA 19104

This research examines how well large language models (LLMs) generate new product ideas, focusing on new products for college students priced under $50. Through a series of studies, we identify key strengths and weaknesses of using LLMs for product innovation. Our first study shows that LLM-generated product ideas have higher average quality than human ideas (based on purchase intent) and are 7 times more likely to rank in the top 10%. Our second study demonstrates that this AI-induced creativity boost is not explained by the LLM's more persuasive pitching skills. Our third and fourth studies identify an interesting weakness of using LLMs for brainstorming, showing that AI-generated ideas are less novel (at the idea level) and less diverse (at the set level). In our fifth study, we investigate the generalizability of these findings by analyzing prior LLM-based creativity studies, finding consistently lower idea diversity across all of them. Our sixth and seventh studies investigate the efficacy of various techniques to mitigate this diversity loss. Specifically, we compare different LLMs (different vendors and different versions) and find that the more recent models are capable of generating more diverse ideas though still falling short of achieving human-level diversity. In addition, we demonstrate the effectiveness of several techniques that can be used to increase idea diversity almost to the level of human idea generation: pooling ideas across vendors, prompt engineering (including Chain-of-Thought prompting and injecting heterogeneous personas or constraints) and creative agents that broadly explore the solution landscape with the intent of restoring diversity. Finally, in our eighth study we show that exploiting the near-zero marginal cost of AI idea generation by scaling the number of ideas steadily improves coverage of the idea space approaching human-level coverage. We conclude our work by presenting a set of actionable recommendations targeted to managers of innovation that want to identify better new product ideas with the help of LLMs.

## 1. Introduction

The dream of using artificial intelligence (AI) for creative work is almost as old as the field of AI itself. Herbert Simon and Allen Newell developed their General Problem Solver, a computer program designed to emulate human problem-solving processes, many decades ago (Newell et al. 1959) and in his essay on creativity, Simon showed a strong conviction that computers were able to engage in creative tasks ("a computer, though it is a machine, need not behave in a machine-like manner"; Simon 1967). In the subsequent decades, a number of creative AI tools were developed, including Pix2pix and CycleGAN for art, WaveNet and NSynth in music, and JAPE for humor (for a historical review see Miller 2019). Despite these early proofs of concept, AI-based creativity remained a niche application with few users and limited economic importance.

In the early 2020s, text-to-image generators such as OpenAI's DALL-E or Stable Diffusion were launched and chatbots, based on the generative pre-trained transformer (GPT) series of large language models (LLMs), such as OpenAI's GPT-3.5 or Anthropic's Claude reached large audiences. These releases pushed model capabilities to previously unseen levels while dramatically simplifying ease of use. By the end of 2022, ChatGPT reached over 100 million users and creative problem-solving tasks including generating ideas and recommendations in various domains, developing software code, and creative writing and poetry emerged among the most common applications (Bubeck et al. 2023; Doshi and Hauser 2024).

With the rapid expansion of Generative AI (GenAI) use cases, a new stream of research has also emerged. In the tradition of the famous Turing Test (Turing 1950), researchers have been recruiting human judges, ranging from laypersons to domain experts, to compare the output of workflows performed by human workers with that of GenAI models. Within months, studies established that GPT and similar models reach human-level competence at tasks such as medical diagnostics (Ayers et al. 2023), persuasion (Costello et al. 2024; Redden 2025) and even solving legal problems (Katz et al. 2024). Such "humanity vs machine" comparisons have also been conducted in the domain of creativity and innovation. For example, a pioneering study in this tradition is Dell'Acqua et al. (2026), who conducted one of the first rigorous human-AI comparisons in a professional consulting context, showing that consultants using AI significantly outperformed those working without it. Further, in analyzing how human-guided AI models perform on a creative challenge, Boussioux et al. (2024) find that LLMs identified solutions of higher strategic viability compared to humans, whereas Lee and Chung (2024) report that ChatGPT is highly effective at solving everyday creative problems yet struggles with creating radical innovation.

The current research follows this paradigm of comparing human-generated work with the output of GenAI models but seeks to move deeper to understand the fundamental trade-offs of using AI for innovation. Specifically, we look at the domain of entrepreneurial idea generation and compare ideas for new products targeted toward college students that were generated by humans with ideas generated by GPT-

4. Using individual ideas as the unit of analysis, we define an idea as a novel match between a solution and a need, and idea generation as a search process aimed at discovering high-value ideas. The significant cost reduction of generating ideas with LLMs is self-evident and not explored here. Instead, we structure our investigation around three central themes.

First, we investigate **idea quality**. We extend the prior work by focusing on the domain of creating an idea for a new product, a setting where success is driven not by the average quality of ideas, but by the quality of the best idea (Girotra et al. 2010; Wooten and Ulrich 2017). This means generating a large quantity of ideas and seeking variance in outcomes are desirable strategies. LLMs may be uniquely suited for this, as their ability to generate a vast number of ideas ("productivity") from a wide range of concepts ("semantic breadth") increases the probability of discovering a high-value outlier (De Freitas et al. 2025). Indeed, previous work has demonstrated that LLMs increase the average quality of ideas in various forms of creative challenges (Hubert et al. 2024; Boussioux et al. 2024; Lee and Chung 2024). Similar to these studies, our comparison includes the average idea quality, but it is the quality of the best idea(s) that we are most interested in. To that end, our Study 1a directly compares the quality of human- and AI-generated product ideas using customer purchase intent, a market-validated metric (Ulrich and Eppinger 2007). Our results for this study show that AI-generated ideas are, on average, of higher quality than human-generated ideas. Beyond this improvement in average idea quality, we find that LLM-generated ideas are 7x more likely to be evaluated as being among the best 10% compared to human-generated ideas. We then address two possible confounders in AI-enabled brainstorming. First, given that AI has been shown to be persuasive (Herbold et al. 2023; Matz et al. 2024; Salvi et al. 2025; Bai et al. 2025) and that pitching is an important part of a product presentation (Kalvapalle et al. 2024), the quality of AI ideas might be driven by superior "pitching skills". However, as we show in Study 1b, AI's superior idea performance cannot be attributed to its pitching or communication skills. Specifically, when human ideas were rephrased by AI to match AI writing style, there was no meaningful difference in purchase intent between original and rephrased ideas. Second, another potential confounder is idea novelty. Given its training process (McCoy et al. 2023; Nguyen 2024), the AI might simply be regurgitating existing ideas from its training data, resulting in less novel ideas. In Study 1c, we show that the idea novelty, as rated by human judges, only decreases by a small amount, suggesting that AI ideas are not merely regurgitations of existing ideas.

Second, we turn to **idea diversity**, which we operationalize as the semantic distance between ideas in a set, following Johnson et al. (2021). Diversity is important for building robust innovation portfolios and ensuring the entire opportunity landscape is explored (Wooten and Ulrich 2017; Kornish and Hutchison-Krupat 2017; Johnson et al. 2021; Si et al. 2022). Research estimating the size of opportunity spaces suggests these landscapes contain thousands of unique ideas (Kornish and Ulrich 2011), underscoring the importance of broad exploration. While previous research has shown that LLMs can create high-quality ideas (Boussioux et al. 2024; Lee and Chung 2024), further research is needed to investigate how AI affects the collective diversity of creative output (Lee and Chung 2025). Here, we anticipate a key weakness of AI based on early scientific evidence (Meincke et al. 2025) as well as the AI training process, which rewards probable associations. This reward system might make LLMs converge on a narrower set of familiar concepts, leading to a portfolio of ideas that are more similar to each other (Doshi and Hauser 2024; De Freitas et al. 2025). This pattern echoes findings from human ideation research showing that without appropriate incentive structures, ideators tend to exploit familiar approaches rather than explore novel ones (Toubia 2006) and that repeated practice of creative tasks can actually narrow cognitive activation, reducing idea novelty over time (Brucks and Huang 2020). Study 2a tests this possibility by comparing the diversity of our human- and AI-generated ideas from Study 1. We show that idea diversity in AI-generated idea sets is substantially reduced (in the sense that ideas are more similar to each other). As a result, the underlying solution landscape is less likely to be fully explored by AI, which is an important requirement in situations of substantial complexity (Sommer and Loch 2004). To examine the generalizability of our findings, Study 2b conducts a meta-analysis, applying multiple diversity metrics to additional data from three recent studies on AI and creativity (Stevenson et al. 2022; Hubert et al. 2024; Boussioux et al. 2024). Across studies and alternative measures of diversity, we find that AI-generated ideas were significantly less diverse than human-generated ideas with effect sizes ranging from small to very large.

Third, we explore **diversification strategies**, i.e., managerial interventions that aim to at least partially restore the AI-induced diversity loss. If AI's tendency is to narrow the search, can we program it to think more broadly? Study 3a investigates technical levers, such as choosing newer models and adjusting model parameters (e.g., temperature). In addition, we investigate pooling ideas from different model vendors to restore diversity, a strategy akin to open innovation tournaments in human-centered brainstorming (Terwiesch and Ulrich 2009), where parallel search across independent ideators yields limited redundancy (Kornish and Ulrich 2011), and hackathons, where diverse teams working in parallel on compressed timelines can produce breakthrough innovations (Lifshitz-Assaf et al. 2021). We find that the more recent models are capable of generating more diverse ideas and that one can boost diversity by pooling ideas from different model vendors. Study 3b focuses on procedural solutions, testing whether advanced prompt engineering (e.g., Chain-of-Thought, Wei et al. 2023) and the use of specialized "creative agents" working alongside human ideators can counteract AI's homogenizing tendency and restore the diversity lost. We show that Chain-of-Thought prompting is effective, as is injecting heterogeneous elements—such as randomly assigned personas or functional design constraints—into the generation process. We also find that a creative agent that uses AI to generate solutions to open needs identified by humans is particularly successful at increasing idea diversity. In addition, Study 3c shows that managers can leverage AI's cost advantage by generating large and diverse idea pools: as the number of AI ideas increases from 100 to 1,000, the idea space is covered more and more completely without plateauing, almost reaching human-level coverage, though some human-explored regions remain difficult for AI to reach.

We conclude by presenting a set of actionable recommendations targeted at managers of innovation who want to use AI to identify better ideas for future new products, discuss the limitations of our findings, and propose an agenda for further research on the topic. Table S1 provides a detailed summary of our research aims and contributions.

## 2. Study 1: The Impact of AI on Idea Quality

Our first study focuses on comparing the quality of the AI-generated to human-generated ideas. To illustrate our unit analysis of an idea, consider this student-generated idea from a class taught by one of the authors:

*Convertible High-Heel Shoe: Many prefer high-heel shoes for dress-up occasions, yet walking in high heels for more than short distances is very challenging. Might we create a stylish high-heel shoe that easily adapts to a comfortable walking configuration, say by folding down or removing a heel portion of the shoe?*

In this example, the need is some people's desire to dress up and wear high-heeled shoes for some occasions, while still walking comfortably. The proposed solution is to make the heel portion of the shoe so that it can be folded down or removed.

Idea generation, in general, and brainstorming in particular, is a process that creates a stream of ideas. This stream can be the result of human effort and/or the use of AI. Generating an idea has a cost. While this cost might be small, it is not negligible, especially if ideas are generated in corporate settings (as opposed to everyday activities, like coming up with the idea for a birthday gift). In managerial settings, after factoring in meeting time, travel expenses, and overhead costs, standard cost estimates range between tens of dollars and hundreds of dollars per idea (Terwiesch and Ulrich 2009). The most obvious and dramatic effect of using LLMs for brainstorming purposes is that the cost of generating one additional idea can be dramatically reduced. However, as stated above, the effect of the unit cost reduction is so significant and self-evident that we will not explore it further in this paper.

In addition to the cost, a second important attribute of an idea is its quality. Some ideas are outright horrible (very low quality), a few are outstanding (very high quality), and most fall somewhere in between. At the moment of idea generation, the quality of the next idea that arises is typically unknown and thus can be thought of as a draw from an underlying probability distribution (Weitzman 1979). This distribution is typically referred to as the parent distribution (Dahan and Mendelson 2001).

The idea-generation process can thereby be thought of as a search process that generates ideas with random quality values by drawing from an underlying stochastic distribution. Prior work has modeled this search as involving trade-offs between exploiting known fruitful approaches and exploring uncertain alternatives (Toubia 2006). Note that the overall success of the idea generation in a new product development setting like ours does not depend on the *average* quality of the idea, but is driven by the quality of the *best* idea (Dahan and Mendelson 2001; Terwiesch and Ulrich 2009; Girotra et al. 2010). Moreover, similar to financial options, ideas have an optionality built into them. Ideas can never have a negative value, as low-quality ideas can always be discarded. For that reason, unlike in most settings of operations management, in innovation, "variance is your friend" and the innovating manager should seek high variance pay-offs as opposed to being risk-averse. As argued by Girotra et al. (2010), the quality of the best idea thus depends on (1) the total number of ideas generated (more ideas lead to a better best idea), (2) the average idea quality of the parent distribution (on average better ideas lead to a better best idea), and (3) the variance of the parent distribution (higher variance ideas lead to a better best idea).

Given this search-based view of ideation, recent theoretical insights suggest that LLMs may be uniquely suited to support the search for ideas of exceptional quality. De Freitas et al. (2025) propose that LLMs mirror two core psychological pathways observed in human creativity: productivity, or the ability to generate a large number of ideas quickly (akin to persistence), and semantic breadth, or the ability to draw from a wide range of conceptual categories (akin to cognitive flexibility). These two properties directly support the two levers of innovation success: generating more ideas and increasing their quality. LLMs are particularly efficient at high-throughput ideation, producing hundreds of candidate ideas without fatigue or coordination costs, and their training on vast and heterogeneous corpora enables associative jumps across otherwise distant semantic domains, driving higher variance and increasing the chances of sampling extreme, high-impact ideas.

Crucially, LLMs may also contribute positively to the average quality of ideas. Because they have been trained on trillions of tokens of human-generated content, including high-quality writing, product descriptions, and creative analogies, they tend to produce outputs that are fluent, coherent, and pragmatically appropriate. This allows them to generate many ideas that are not only original but also feasible and well-articulated, particularly in domains where quality is partly defined by communication effectiveness or alignment with existing cultural references.

The first aim of our research is to compare the quality of human-generated ideas with the quality of AI-generated ideas. Research to date has demonstrated that AI frequently matches or exceeds human performance in creative tasks. Haase and Hanel (2023) found that LLMs reached human-level performance in divergent thinking tasks such as the Alternative Uses Task (AUT). This is supported by Hubert et al. (2024), who studied GPT-4 response performance in the Consequences Task (CT) and Divergent Association Tasks (DAT), finding that AI is more creative than humans across all its dimensions. However, there is also some literature showing the primacy of humans, or humans plus AI, in certain contexts. For example, Koivisto and Grassini (2023) find that while AI chatbots outperform average human performance in the AUT, the most exceptional human ideas still match or exceed those generated by AI. Lee and Chung (2024) report that humans using ChatGPT generate more creative ideas than humans using traditional web search or than humans working alone, across small creative problems such as repurposing an item or coming up with an idea for a gift. Closest to the current study, Boussioux et al. (2024) analyze how human-guided AI models perform on a creative challenge of generating business ideas for a circular economy, specifically through a crowdsourcing contest that asked participants to propose sustainable business opportunities emphasizing reuse, regeneration, and circular design principles. They find that AI-generated solutions exhibit higher strategic viability than human-generated ones, but score lower on novelty. Joosten et al. (2024) show that AI-generated ideas score significantly higher in novelty and customer benefit in a blind evaluation comparing human professionals and ChatGPT as ideators tasked with generating innovative and sustainable packaging solutions.

We add to this line of inquiry in our Study 1a by (1) focusing on the domain of generating ideas for a new product, which is characterized by a payoff that is driven by the best idea(s) as opposed to relying on the average idea quality, (2) utilizing market research data of real customers in assessing the quality of the idea and (3) comparing the difference between zero-shot and few-shot prompting.

As we evaluate idea quality, two concerns arise. First, it is important to disentangle the quality of the idea itself (the "raw" idea quality) from the quality of the idea articulation (how the idea is "pitched"). In the context of pitching business ideas to seek funding, an extensive literature in the field of Entrepreneurship has demonstrated the importance of good pitching skills (for a review see Kalvapalle et al. 2024). Similar findings have been demonstrated in the context of research grant proposals (Peng et al. 2024). Relatedly, previous research has shown that LLMs outperform humans in persuasion across many tasks, including articulating ideas more clearly (Herbold et al. 2023; Matz et al. 2024; Salvi et al. 2025; Bai et al. 2025), which might lead them to craft especially convincing product pitches. Study 1b seeks to disentangle the quality of the pitch from the quality of the idea.

Second, in the context of a human-to-AI comparison of idea quality, we acknowledge an important advantage of AI. The corpus of data that the LLMs have been trained on includes product catalogs and assortments provided by (online) retailers. Previous work has shown that LLMs can reproduce verbatim text snippets from their training data (McCoy et al. 2023; Nguyen 2024), raising concerns about the LLM retrieving existing ideas and presenting them as novel. In their idea generation of strategies for a circular economy, Boussioux et al. (2024) ask human judges to assess the novelty of the generated solutions and find that the use of AI decreased novelty. In contrast, Joosten et al. (2024) report an increase in idea novelty as assessed by one human expert evaluator. To address this tension, Study 1c evaluates the novelty of the ideas by asking potential customers to evaluate the novelty of AI-generated and human-generated ideas.

### 2.1. Research Setting

The product ideation task of this study is based on human data collected during a product design and innovation course taught as a part of an MBA program at a US university in 2021, before the widespread adoption of LLMs. In this course, 50 students participated in an innovation challenge to come up with ideas for a physical product targeted towards college students that could be sold for $50 or less. This price cap was imposed to limit the complexity of the corresponding product development projects to what could reasonably be expected from student teams to be accomplished in a one-semester course. The innovation challenge was organized in a traditional innovation tournament format (Terwiesch and Ulrich 2009), where individuals first independently generate many ideas, which are then combined into a pool of several hundred ideas and subsequently evaluated by others in the group (i.e., "crowdsourced" evaluations). Thus, we have access to a large set of ideas generated by humans shortly before AI tools became widely available to enhance ideation. From this large pool of independently aggregated human ideas, we randomly select 200 entries to assess their quality and novelty.

In our first study, we include 200 AI-generated ideas across two pools of 100 ideas each. We generated ideas for the first sub-pool via "zero-shot prompting". That is, we prompted OpenAI's GPT-4 (more specifically, gpt-4-0314) with the same prompt we gave the students and without any references to past ideas. We generated the second pool of 100 ideas via "few-shot prompting" providing the model with six ideas that scored well in prior offerings of the class (details about the exact prompting protocols can be found in E-Companion A). This approach allows the model to potentially understand what features high-quality ideas exhibit.

The resulting average word counts for GPT-4-generated ideas with zero- and few-shot prompting were 69 and 71 respectively, versus 63 words for student ideas. To ensure that the few-shot prompted ideas produced by GPT-4 were not merely reused or slightly edited versions of the initial prompts, we manually reviewed them and confirmed their difference. We generated our initial two GPT-4 pools (zero-shot and few-shot) using a temperature of 0.7 (a parameter controlling how deterministic output is, where 0 is close to deterministic and higher values indicate greater randomness), which was the default at the time of the experiments in May 2023.

The research described in this paper was approved by the Institutional Review Board (IRB) at our home institution under Protocol #853634.

## 2.2. Study 1a: The Quality of AI-generated ideas

**Method:** We compare the previously defined three pools of ideas (200 human-generated ideas, versus 100 AI-generated ideas with zero-shot prompting and 100 AI-generated ideas with few-shot prompting) according to their idea quality. To obtain an independent measure for idea quality, we used the online platform Prolific to recruit college-age individuals from the United States to evaluate all 400 ideas via a purchase-intent survey. While recent work has demonstrated that AI agents can infiltrate online surveys and behavioral experiments (Westwood 2025; Van der Stigchel et al. 2026), empirical audits of Prolific — the platform used here — estimate AI contamination at approximately 4% (Westwood and Frederick 2026; Chen et al. 2026) and potentially below 1% (Gordon et al. 2026). Kornish and Ulrich (2014) show that the best indicator of future value creation is the average purchase intent expressed by a sample of consumers in the target market. Specifically, we asked respondents to express their purchase intent using the standard five-box options: “definitely would not purchase”, “probably would not purchase”, “might or might not purchase”, “probably would purchase”, and “definitely would purchase”. To translate these responses to probabilities, we follow one of the weighting schemes used by Jamieson and Bass (1989) with the corresponding weights [0, 0.25, 0.50, 0.75, and 1.00], and report robustness tests using other weighting schemes. Each respondent evaluated an average of 40 ideas with randomized source and order. Each idea was evaluated by 20 participants on average.

We are interested in the average idea quality of the three pools as well as the proportion of the top 10% of ideas that come from each of the three pools. We formally tested the impact of idea source on the perceived quality of product ideas via a linear mixed-effects model with purchase intent as the dependent variable. The model included two fixed-effects denoting source (humans are the baseline) and random intercepts and slopes for respondents and ideas, allowing source effects to vary across evaluators and ideas. We provide additional robustness checks by controlling for word count.

To better understand how the full distribution of idea qualities is affected by the idea source, we use quantile regression analysis (Koenker and Hallock 2001). Quantile regression extends traditional regression by computing the relationship between explanatory variables (e.g., idea source) and the response variable (e.g., idea quality), examining whether the effect of AI versus human generation differs between percentiles (e.g., at the top 10% of all ideas versus the top 50% of all ideas). Using quantile regression, we can examine the tails of the distribution instead of the mean, allowing us to test whether GPT-4 excels at generating high-quality ideas only for specific percentiles — such as producing more ideas at the upper percentiles — or whether the effect holds across the entire distribution. Following Girotra et al. (2010), we use the average idea quality ratings as the dependent variable, and our explanatory variable is a binary indicator for whether the idea is human-generated or AI-generated.

**Results:** Figure 1 shows the quality (purchase probability) of ideas across the three pools. On average, GPT-4-generated ideas had greater purchase intent (46.4% with zero-shot prompting and 49.3% with few-shot prompting) than humans (40.4%). The standard deviation of idea quality was comparable between the three pools. We found significant differences in the perceived quality of ideas as a function of their source.

Ideas generated by GPT-4 with zero-shot prompting were rated significantly higher than human-generated ideas (*B* = 0.059; 95% CI [0.030, 0.088]; $t(246) = 4.06$, $p < .001$; see Table S2) and ideas generated by GPT-4 with few-shot prompting received even higher ratings (*B* = 0.089; 95% CI [0.060, 0.119]; $t(223) = 5.93$, $p < .001$; see Table S2). The difference in purchase intent between the two AI-generated idea pools was substantially smaller in magnitude and did not reach statistical significance (*B* = 0.030; 95% CI [-0.001, 0.060]; $t(200) = 1.90$, $p = .057$; see Table S3). These findings indicate that GPT-4-generated ideas are, on average, more likely to be purchased than human-generated ideas. See Tables S4, S5 and S6 for robustness checks using an alternative weighting scheme (Ulrich and Eppinger 2007), a simpler model and controlling for word count (all with similar results).

Beyond averages, we find that the benefit of using GPT-4 for idea generation becomes even more substantial at the tails of the quality distributions. For all percentiles, GPT-4 ideas consistently outperformed student ideas (see Figure S1 and Table S11 for all results).

**Figure 1.** Distribution of Idea Quality for Three Sets of Ideas

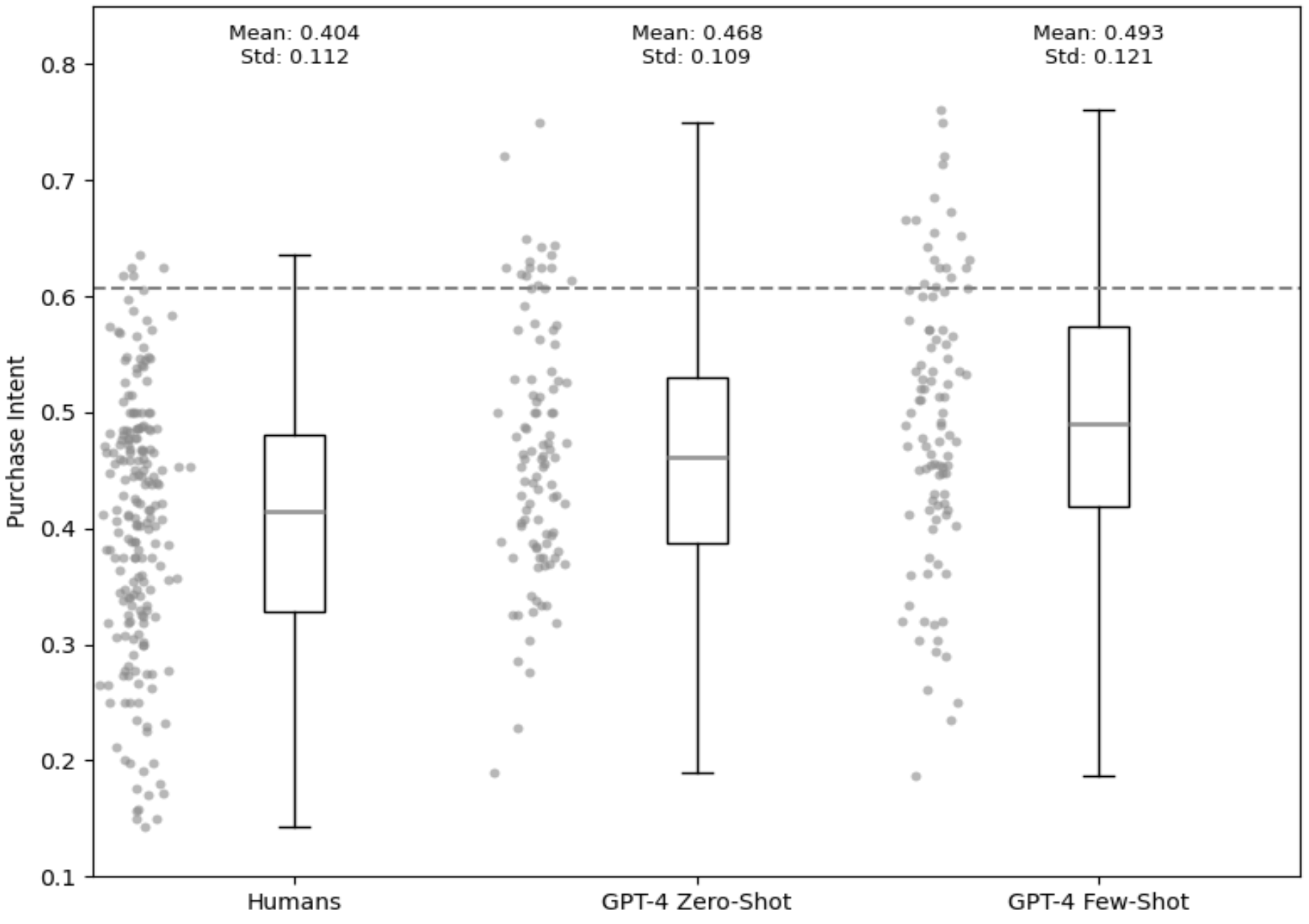

*Notes.* Each dot represents one idea's purchase intent rating averaged across individual ratings. Horizontal straight line indicates mean rating. Ideas above the horizontal dashed line are the top 10% ideas. Purchase intent is the average of the five-box response scale per Jamieson and Bass (1989).

The effect was especially pronounced for the upper tail of the distribution (80% and above), where GPT-4 had the strongest advantage. Among the upper 10% of the ideas in terms of quality (i.e., 40 out of the 400), only 5 came from humans, compared to 15 ideas from the zero-shot-prompted LLM pool, and 20 from the few-shot-prompted LLM pool.

## 2.3. Study 1b: Disentangling Idea Quality from Pitching Skills

**Method:** We use our sample of 200 human ideas and prompt an LLM (gpt-4-0613) to rephrase each human idea in a style mimicking the LLM-generated idea from Study 1a (see E-Companion B for full prompt). Similar to Study 1a, the AI-generated ideas have a slightly greater word count (81.61, SD: 18.63) than the human ones (62.71, SD: 34.41). We again formally tested the impact of idea source on the perceived quality of product ideas via a linear mixed-effects model with purchase intent as the dependent variable and random intercepts and slopes for ideas and raters, accounting for the pairing of original and rewritten ideas. As robustness, we conducted a paired t-test on product-level mean ratings. For the ratings, we recruited a new human panel from Prolific where each rater evaluated 40 ideas (20 original, 20 rephrased). No rater saw both versions of the same idea, ensuring between-subjects independence at the rater-idea level.

**Results:** Rephrasing human ideas using AI did not significantly change their perceived quality, with the 95% CI excluding effect sizes larger than 2.3 percentage points ($B = 0.009$; 95% CI [-0.005, 0.023]; $t(142) = 1.27$, $p = .205$; see Table S7). Results are robust to controlling for word count (see Table S8). A quantile regression did not show significant rephrasing effects for any specific quantile (see Figure S2), suggesting that the previously observed advantage of GPT-4-produced ideas in the top decile is not attributable to writing style and does not affect ideas at different quantiles differently. The ratings between original and rephrased ideas are significantly correlated $r(200) = 0.53$ (95% CI [0.43, 0.63]; $p < .001$). This t-test confirms the mixed model result with a small and not significant quality difference ($\Delta = 0.009$, $t(199) = 1.26$, $p = .208$).

## 2.4. Study 1c: The Impact of GenAI on Idea Novelty

**Method:** We surveyed an additional panel of human raters on Prolific (akin to Studies 1a and 1b, with 40 ratings per respondent, randomized and blinded, 20 ratings per idea) on idea novelty. Following Shibayama et al. (2021), we measured perceived novelty using a five-point scale where evaluators rated how novel each idea appeared "relative to other products you have seen" (0: Not at all novel, 0.25: Slightly novel, 0.5: Moderately novel, 0.75: Very novel, 1: Extremely novel). We employed linear mixed-effects models as our primary analytical approach, accounting for the nested structure of our data where multiple evaluations are clustered within ideas and evaluators. Models included fixed effects for idea source (human vs. AI conditions), random intercepts and slopes for individual evaluator and idea-level variation, allowing source effects to vary across evaluators and ideas.

**Results:** The average novelty of human-generated ideas was 40.6% (SD: 0.117), greater than that of zero-shot GPT-4 (36.7%, SD: 0.101), and few-shot GPT-4 (36.1%, SD: 0.11). We found significant differences in perceived novelty between human and both zero-shot GPT-4-generated ideas ($B = -0.038$; 95% CI [- 0.066, -0.01]; $t(269) = -2.67$, $p = .008$; see Table S9) and few-shot GPT-4 ideas ($B = -0.049$; 95% CI [-0.078, -0.02]; $t(268) = -3.40$, $p < .001$; see Table S9). Perceived novelty was not significantly different between the two pools of LLM-generated ideas ($B = -0.011$; 95% CI [-0.039, 0.017]; $t(186) = -0.79$, $p = .431$; see Table S10). Novelty correlated negatively with purchase intent, but the correlation magnitude was small and not statistically significant $r(400) = -0.078$ (95% CI [-0.174, 0.021]; $p = .124$).

### 2.5. Discussion

Using purchase intent as our measure of idea quality, we replicate the finding of prior studies that LLMs are capable of performing well at creative tasks, in this case producing high-quality ideas. When focusing on the quality of the best ideas the LLM performance looks even more promising, with GPT-4-generated ideas being 7x more likely to appear among the top 10% of ideas compared to human ideas. Thus, not only does GPT-4 generate better ideas on average, but it is also especially capable of producing top-tier ideas. We also found that employing few-shot prompting, i.e., including successful ideas from the past in the prompt, improves the average idea quality as well as the quality of the best ideas.

Our results further indicate that the superior performance of AI-generated ideas cannot be attributed merely to GPT-4's persuasive pitching skills. The quality advantage documented in Study 1a thus reflects a genuine difference in the underlying idea quality regardless of the communication. One plausible explanation for this result is specific to our students, who have gone through a rather selective admissions process that likely provides an edge to good communicators and who also have received substantial training related to communications and pitching. In other words, we cannot rule out the possibility that our human subjects were exceptionally persuasive salespeople and GPT-4 might retain a pitching advantage over laypeople.

Despite not explicitly prompting the LLM for creative or novel ideas, the produced ideas were only slightly less novel than human ideas, suggesting that it does not overly rely on existing product ideas but instead creates novel products and recombinations. The decrease in novelty was more pronounced when using few-shot prompting, suggesting that providing examples may have inadvertently anchored the model more strongly to familiar patterns. Our findings are more consistent with the prior work by Boussioux et al. (2024) rather than with what Joosten et al. (2024) report.

## 3. Study 2: The Impact of AI on Idea Diversity

Given that an idea is still far from implementation, the true payoff associated with it is clouded by uncertainty. Therefore, innovation managers typically move beyond selecting individual ideas and instead try to design portfolios of ideas (Si et al. 2022). Ideally, ideas in a portfolio are diverse, meaning that they are not too similar to one another, and cover a large solution space. Johnson et al. (2021) operationalize this idea diversity as the semantic distance between the ideas in a set and we follow this definition.

The importance of idea diversity extends beyond simple risk hedging. Research on human creativity has established that diversity serves as a critical mechanism for enhancing innovation outcomes (Wooten and Ulrich 2017). Specifically, studies show that exposing individuals to a diverse set of initial ideas or examples improves the novelty of the ideas they subsequently generate (Rietzschel et al. 2007; Baruah and Paulus 2011). In organizational settings, idea diversity is linked to increased profitability, as it ensures a firm is not overinvesting in a narrow region of the opportunity space (Gielnik et al. 2012). Therefore, diversity is not merely a passive portfolio strategy but an active ingredient that fuels the creative process itself.

Similar to a potential reduction in novelty, as discussed in Study 1c, it is plausible that LLMs generate portfolios that are less diverse than human-generated ones. The stochastic nature of LLMs, combined with a training process that explicitly penalizes deviation from expected outputs, might prevent them from exploring a broader region of the idea space and instead favor familiar associations and established concepts over unexpected or unconventional combinations (Kirk et al. 2024; Bronnec et al. 2024), an essential ingredient in creative ideation. In humans, Brucks and Huang (2020) show that repeated practice reinforces familiar cognitive pathways while inhibiting less-traveled ones, reducing the novelty of generated ideas. This architectural constraint means that while individual LLM-generated ideas may achieve high-quality ratings, the underlying token-generation process may inadvertently reduce portfolio diversity because LLMs mirror a narrow reward function, which amplifies homogeneity (De Freitas et al. 2025). The implications extend beyond individual innovators: if organizations increasingly rely on AI for ideation, we may observe a form of collective convergence where individual productivity gains come at the cost of reduced aggregate diversity across the innovation ecosystem (Doshi and Hauser 2024; Meincke et al. 2025). This risk is particularly acute given evidence that experienced innovation practitioners intuitively recognize diversity's importance (Johnson et al. 2021), yet automated systems may not incorporate this hard-won wisdom.

To operationalize idea diversity, we first define its converse, which is idea similarity. LLMs represent individual ideas (specifically, their text description) using embedding vectors as numerical representations (Berger et al. 2022; Dell'Acqua et al. 2026). These vectors capture the semantic meaning of individual ideas. Individual ideas can thus be thought of as geometric positions in a highly-dimensional space. This space is also referred to as the idea space. Figure 2 visualizes a projection of the position of all ideas (human and AI) used in Study 1 into a two-dimensional space. We define two ideas to be similar to each other if they are positioned close to each other in the idea space and thus their embedding vectors are pointing in a similar direction (cosine similarity). Thus, while novelty is measured by asking participants how similar an idea is compared to other products they know, similarity measures how similar two or more products are within a set of ideas, irrespective of how novel they may be compared to other existing products.

To illustrate, consider the following three ideas and their positions in Figure 2 (see E-Companion C for the full idea descriptions):

- Idea A ("BookBuddy"): an adjustable book holder that makes reading more ergonomic
- Idea B ("Adjustable Laptop Riser"): a laptop riser that puts laptops into a comfortable viewing angle
- Idea C ("Hydration Harness"): a strap that connects water bottles to backpacks

**Figure 2.** Projection of the 512-Dimensional Idea Space into a Two-Dimensional Space

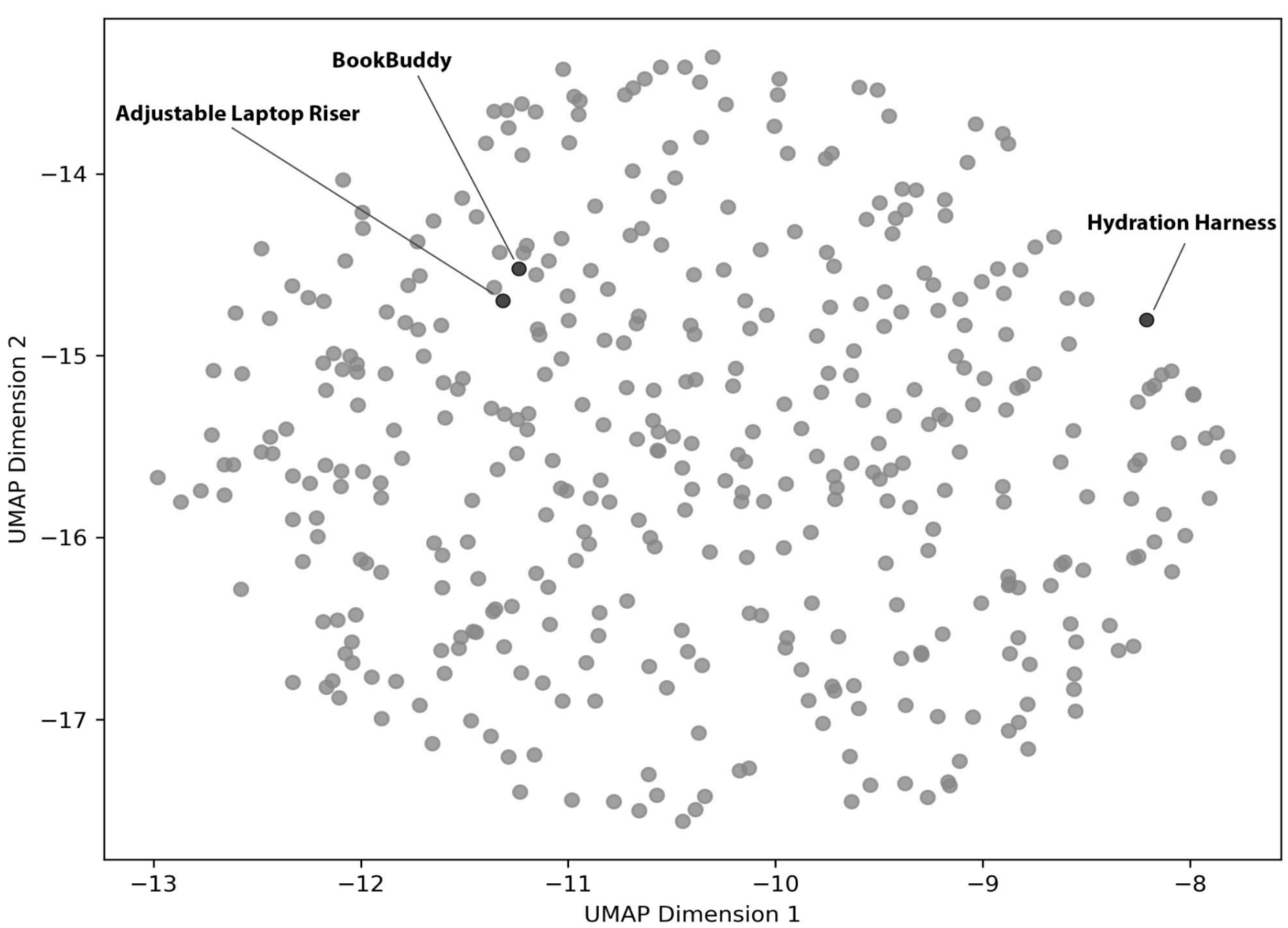

*Notes.* This figure shows the 400 ideas projected from their original USE-based embedding (512-dimensions) down to two-dimensional UMAP (Uniform Manifold Approximation and Projection) space. See E-Companion D for more details.

Both ideas A and B deal with the ergonomic challenges that students have while reading, and both propose a type of adjustable stand as a solution. The ideas are thus similar to each other as captured by their neighboring position in the two-dimensional projection of the idea space in Figure 2. Idea C, on the other hand, is far away from A and B because it aims to solve a completely different problem (student hydration).

In its simplest form we define a set of ideas that are pairwise dissimilar to each other as diverse (Dell'Acqua et al. 2026; Meincke et al. 2025). Several other measures of idea diversity exist, including pairwise distance to an idea set's centroid and a minimum spanning tree (MST; Siangliulue et al. 2016; Cox et al. 2021; Doshi and Hauser 2024) and will be discussed below. Yet, all are built on the concept of ideas not being similar to each other. Note that idea quality and idea novelty discussed in Study 1 are attributes of an individual idea, while idea similarity is an attribute of a pair of (two) ideas and idea diversity can be thought of as a property for any number of two or more ideas. Study 2a compares the diversity of human-generated ideas with the diversity of AI-generated ideas.

The emerging evidence that AI reduces diversity of creative output comes with important caveats. Doshi and Hauser (2024) document the quality-diversity tradeoff in a creative writing task, and Meincke et al. (2025) find a similar pattern when reanalyzing the data from Lee and Chung (2024), where participants used ChatGPT for everyday creative challenges such as coming up with gift ideas. However, the tasks in the original study were set in non-managerial, everyday contexts and were designed to assess individual creativity rather than group-level idea diversity, highlighting the need for further research into the conditions under which LLMs affect the diversity of creative output (Lee and Chung 2025).

Study 2b seeks to generalize the findings of Study 2a, by examining four prior studies of AI in brainstorming: (a) our own data, (b) data from Boussioux et al. (2024), that the authors kindly provided, (c) publicly available data from Hubert et al. (2024), and (d) publicly available data from Stevenson et al. (2022). We follow Cox et al. (2021), who convincingly argue that any single measure of idea diversity has its limitations, and recommend considering multiple measures to capture the diversity of a pool of ideas, which we include in Study 2b.

### 3.1. Study 2a: Idea Diversity in the College Products Setting

**Method**: To determine how distinctly different an idea is relative to other ideas, we measured the cosine similarity of its embedding vector relative to the embedding vectors of the other ideas in the same pool (following Cox et al. 2021 and Dell'Acqua et al. 2026). While negative values are possible, they rarely occur in practice. By performing a pairwise comparison of all ideas and averaging their similarities, we computed the average pool similarity.

We further rephrased all 400 ideas using an LLM and compared differences in cosine similarity before and after rephrasing. This allows us to establish whether cosine similarity differences are primarily driven by topical diversity instead of potentially greater semantic variation in human ideas. See E-Companion E for details.

Lastly, we examined how human- and AI-generated ideas differ across the idea space (distance analysis), across joint quality-novelty quadrants (median split), and across the full distributions of each attribute (shift-function analysis).

**Results:** A descriptive visualization of our main result in Study 2a is shown in Figure 3. The left side of the figure shows the idea space for human-generated ideas and the right side shows the one for AI-generated ideas. A simple visual inspection suggests that human-generated ideas seem to be more dispersed compared to AI-generated ideas. To see this, consider that in the 5 x 7 = 35 squares embedding space in Figure 3, 20 squares contain at least one human-generated idea, compared with only 16 for AI-generated ideas.

**Figure 3.** 2D Projection of 512-Dimensional Human (Left) and AI (Right) Ideas

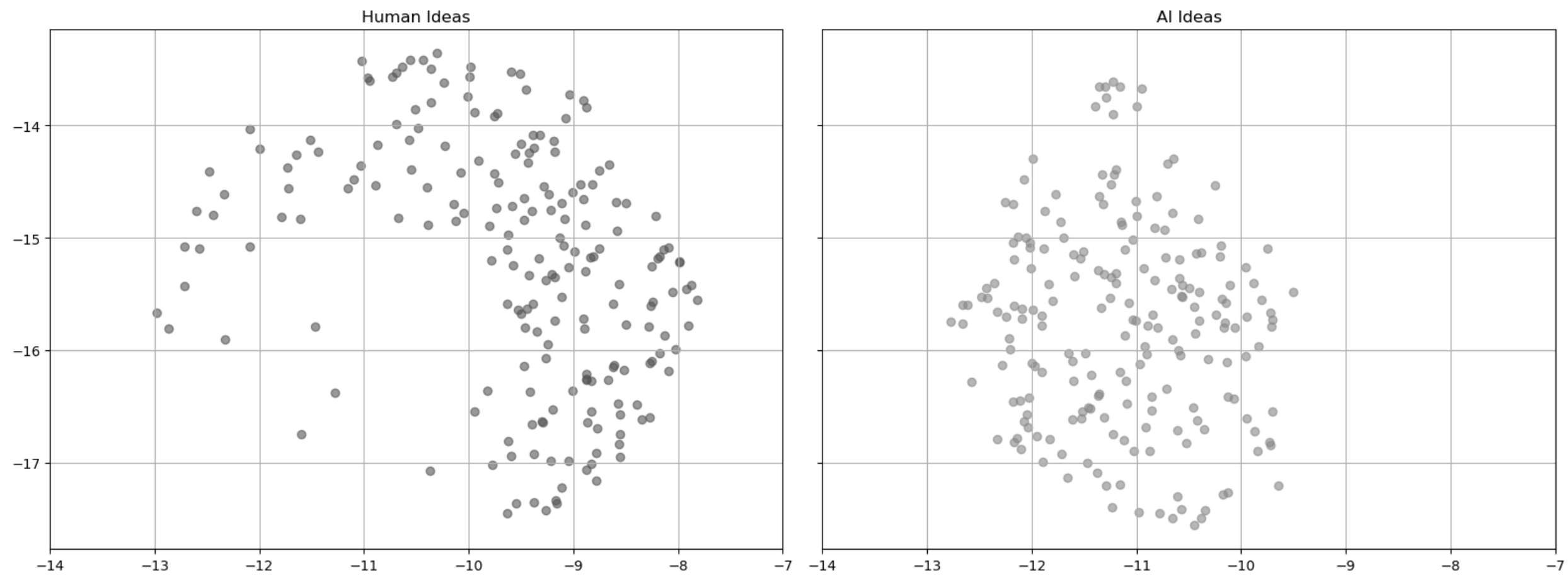

*Notes.* Figure shows 200 human ideas (left) and 200 AI ideas (right, both zero-shot and few-shot). The ideas are projected from their original USE-based embedding (512-dimensions) down to two-dimensional UMAP space. In this figure, the dimensions are shared between both conditions by embedding and projecting down all 400 ideas at once. In all statistical analyses, the pools are analyzed separately.

For each idea, we computed the average pairwise similarity between the idea and all other ideas in the same pool (see Figure 4).

**Figure 4.** Similarity of Ideas Across Human and GPT-4 Pools

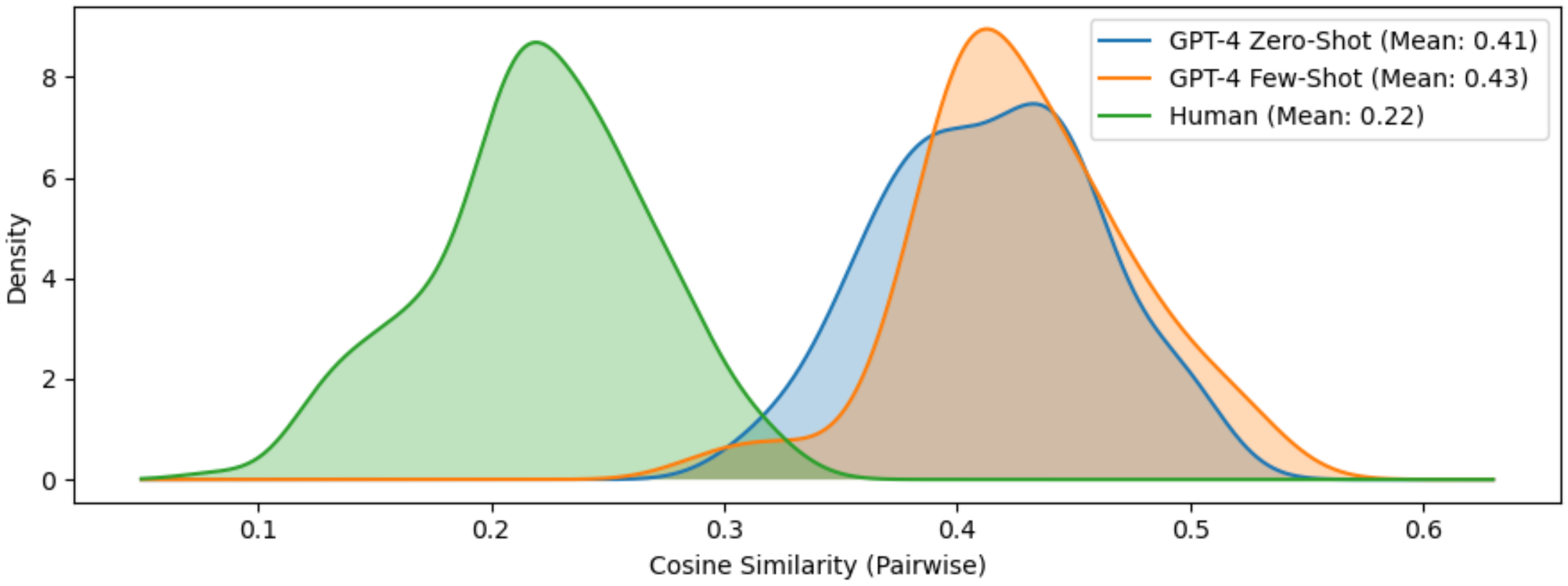

*Notes.* Density plot using pairwise cosine similarity.

A combined model (human, GPT-4 zero- and few-shot) showed a significant overall effect of condition on similarity ($F(2, 397) = 919.6$, $p < .001$, see Table S12, Model 1). Pairwise comparisons revealed that GPT-4 zero-shot ideas were significantly more similar to each other than human outputs ($B$ =

0.193; 95% CI [0.182, 0.204]; $t(298) = 33.89$, $p < .001$, see Table S12, Model 2). Few-shot ideas were less diverse than both human ideas ($B = 0.207$; 95% CI [0.196, 0.219]; $t(298) = 35.98$, $p < .001$, see Table S12, Model 3) and zero-shot ideas ($B = 0.014$; 95% CI [0.001, 0.027]; $t(198) = 2.12$, $p = .036$, see Table S12, Model 4). Results were robust even after rephrasing all ideas, with 80% (zero-shot) and 90% (few-shot) of the diversity gap persisting. This suggests that cosine similarity captures genuine topical/substantive differences rather than merely stylistic differences (see E-Companion E for details and more results).

An OLS regression treating cosine distance as a continuous predictor ($B = -0.288$, 95% CI [-0.424, -0.153], $t(198) = -4.19$, $p < .001$, see Table S13) indicated that the farther a human idea is from the nearest LLM idea, the lower its purchase intent. For novelty, there was no significant effect ($B = -0.078$, 95% CI [-0.226, 0.071], $t(198) = -1.03$, $p = .303$, see Table S13). Using pooled medians to divide the idea space into four quadrants, humans and AI produced exactly the same number of high-purchase-intent/high-novelty ideas, with AI dominating the high quality and low novelty quadrant and humans dominating the high novelty and low-quality quadrant. See Figures S4 and S5 for more details.

### 3.2. Study 2b: Meta-Analysis of GenAI's Effect on Idea Diversity

Boussioux et al. (2024) compared 54 human-generated and 30 GPT-4-generated business ideas focused on sustainable, circular economy opportunities. Their GPT-4-assisted conditions distinguish between independent search (multiple independent generations for one input prompt) and differentiated search (one generation at a time, with human follow-up for a different solution afterwards). Both search conditions use three different prompt-engineering techniques (no special prompt, random persona, expert persona), totaling six conditions overall. The authors have two separate innovation challenges. The first one seeks unmet user needs in the context of a circular economy while the second one seeks solutions to these unmet needs. Thus, we obtain two datasets from their study.

Hubert et al. (2024) experimented with 151 U.S. adults recruited via Prolific, comparing the responses to a creative problem-solving task (consequences task) with responses obtained from GPT-4. Our analysis compares the pool of ideas from their human responses ($N = 1{,}781$) with the pool of GPT-4-generated responses ($N = 1{,}710$).

Stevenson et al. (2022) recruited 42 Dutch students who generated 774 ideas for an alternative use task and compared them with 781 GPT-3-generated ideas.

While data from Haase and Hanel (2023) was publicly available, it only contained one idea per model, making diversity estimation impossible.

**Method:** Since the work by Boussioux et al. (2024) is based on two datasets (problems and solutions), these four studies yield a total of five datasets.

We followed the recommendation of Cox et al. (2021) to employ a variety of methods to measure idea set diversity and used the following three. First, following Dell'Acqua et al. (2026), we conducted a pairwise comparison across all possible idea pairs in the pool and then computed the average pairwise similarity of each idea as measured by the cosine similarity of the respective embeddings. A lower similarity score indicates a more diverse set.

Second, following Doshi and Hauser (2024), we measured each idea's similarity relative to the collective set by computing the cosine similarity between individual idea embeddings and the centroid (mean embedding) of all remaining ideas. The diversity score for each pool was calculated as the average of these similarity values. Again, a lower similarity score indicates a more diverse set of ideas.

Third, following Cox et al. (2021) and Siangliulue et al. (2016), we considered all ideas in the pool and computed the Minimum Spanning Tree (MST) dispersion. An MST connects all ideas while minimizing the sum of the edge distances as given by the respective cosine similarities. We then took the sum of the edge distances and divided it by the total number of edges, to obtain the mean edge length. The higher the average edge length, the more diverse we consider a set.

Lastly, we estimated a mixed-effects model pooling across all five datasets with a fixed effect for source (human = 0, AI = 1) with random intercepts and slopes for each individual study.

**Results:** As Table 1 shows, AI-generated ideas were significantly less diverse than human-generated ideas across all four studies (five datasets) using pairwise cosine similarity. In every comparison, AI systems (GPT-3, GPT-4) produced ideas with higher similarity scores, indicating reduced diversity, with effect sizes ranging from small (Cohen's $d$ = 0.380) to very large ($d$ = 5.538).

Analyzing our data, we found similar results to the diversity analysis in Study 2a (see Table 1 for results). We obtained comparable results using alternative diversity measures, with all AI conditions being significantly less diverse (32/32 comparisons). We provide full results across all comparisons in Tables S14 and S15. Pooling across all five datasets, we find that LLM-generated ideas are significantly less diverse ($B$ = 0.090; 95% CI [0.038, 0.142]; $t$(4.01) = 3.38, $p$ = .028) than human ideas (see Table S16).

**Table 1.** Similarity Analysis Across Four Studies Between Human and LLM-Assisted Generations

| Comparison | Human Group Result | AI-aided Result |
|---|---|---|
| | **Present work - Study 1** | |
| GPT-4 zero-shot | 0.221 | 0.414*** |
| GPT-4 few-shot | 0.221 | 0.428*** |
| | **Boussioux (Problems)** | |
| Single problem | 0.254 | 0.465*** |
| Single problem (random persona) | 0.254 | 0.56*** |
| Single problem (expert persona) | 0.254 | 0.448*** |
| Multiple problems | 0.254 | 0.582*** |
| Multiple problems (random persona) | 0.254 | 0.642*** |
| Multiple problems (expert persona) | 0.254 | 0.568*** |
| | **Boussioux (Solutions)** | |
| Single solution | 0.382 | 0.562*** |
| Single solution (random persona) | 0.382 | 0.62*** |
| Single solution (expert persona) | 0.382 | 0.559*** |
| Multiple solutions | 0.382 | 0.596*** |
| Multiple solutions (random persona) | 0.382 | 0.678*** |
| Multiple solutions (expert persona) | 0.382 | 0.618*** |
| | **Hubert** | |
| GPT-4 | 0.112 | 0.27*** |
| | **Stevenson** | |
| GPT-3 | 0.172 | 0.19*** |

*Notes.* Diversity comparisons using pairwise cosine similarity (method 1). Smaller values indicate greater diversity. *P*-values are based on two-sided *t*-tests and are uncorrected. See Tables S14 and S15 for full results across all metrics. See Table S17 for effect sizes, confidence intervals and test statistics. *** $p < .001$.

**Discussion:** While LLMs can generate high-quality ideas (Study 1), they generate ideas that are less collectively diverse in our analysis. The lower diversity suggests that the tested LLMs tend to explore a more constrained region of the solution landscape, leading to an idea pool that clusters in some regions of the idea space and leaving other territory unexplored, which may hinder finding the best solution (Sommer and Loch 2004). This finding complements the analysis of creative writing tasks presented by Doshi and Hauser (2024) and Meincke et al. (2025). Diversity is not a goal in itself, but rather a means to enhance the quality of the best ideas through broader search, reduced redundancy and an increased likelihood that a portfolio of ideas is robust across different assumptions, use cases, and payoff profiles. It is noteworthy that in our setting LLMs can generate exceptionally strong ideas (as shown in Study 1), even while exhibiting limitations in producing diverse sets of ideas and LLMs do not simply miss good ideas or bypass bad ideas. Rather, they appear to concentrate in regions of the idea space that are commercially plausible but less novel, whereas humans explore more unusual regions that are often lower in purchase intent.
The convergent findings across diversity methods and different creative tasks, different AI models, and various prompting strategies, indicate that reduced idea diversity may be a general property of LLMs.

The magnitude of this effect varies, with effect sizes ranging from small ($d = 0.380$) in Stevenson et al.'s study to very large ($d = 5.538$) in certain conditions of Boussioux et al.'s work. This variation likely reflects differences in task complexity, domain specificity, and the particular creative challenges presented to both human and AI participants. Notably, the effect appears most pronounced in more complex, open-ended ideation tasks such as circular economy problem-solving, where the solution space is vast and multifaceted. While our previous studies demonstrated that AI can generate higher-quality ideas on average and excels at producing top-tier ideas, the reduced exploration of the solution landscape suggests that AI-assisted ideation processes may miss potentially breakthrough innovations that lie in less explored regions of the idea space. This creates a tension: AI provides superior average performance and better top-tier ideas within the regions it explores, but it may systematically overlook areas where revolutionary innovations could emerge. Such a loss of diversity is especially problematic in the contexts where the quality of the best idea determines success.

## 4. Study 3: Increasing Diversity

The third aim of our research is to find methods to counter the diversity-reducing effect of AI-powered idea generation identified in Studies 2a and 2b. The first set of methods (Study 3a) investigates the role of vendor selection, model selection, and model configuration. In addition to OpenAI's GPT, other LLMs like Claude, Gemini, and DeepSeek provide plausible alternatives. GPT-4 outperforms GPT-3.5 on many dimensions. For example, GPT-4 scored in the 90th percentile of the Uniform Bar Exam, while GPT-3.5 only scored in the 10th percentile. In its release notes, OpenAI explicitly states that "GPT-4 is more creative [...] than ever before" (OpenAI 2023). But, does it give everyone the same (good) ideas? Idea diversity metrics provide an objective benchmark that can be applied to different models and configurations. Relatedly, Haase et al. (2025) studied evolution in model creativity (including more recent models such as GPT-4.5, o3-mini and Sonnet 3.7) over the last two years, and found no significant improvements when using the DAT assessment as a benchmark.

Comparing different models is of high managerial relevance as innovation managers need to choose a model and parameter values to configure it. Another practical consideration for innovation managers is whether or not one can increase idea diversity by creating a composite pool of ideas that mixes ideas from different LLMs, akin to how innovation tournaments mix the ideas of human ideators (Terwiesch and Ulrich 2009). Research on innovation contests has shown that observing diverse breakthrough solutions helps participants search solution landscapes differently (Wooten 2022), suggesting that exposing innovation managers to ideas from multiple LLM vendors could similarly broaden their subsequent search. A particularly intriguing concept we investigate is to randomly combine ideas from different models in the hope of increasing diversity, similar to a crowdsourcing or open innovation contest (Boudreau et al. 2011; Poetz and Schreier 2012). Importantly, Kornish and Ulrich (2011) show that parallel independent idea generation produces low redundancy, even for narrowly defined domains, suggesting that pooling ideas across different LLM vendors might similarly yield diverse, non-redundant idea sets. Moreover, the stochastic behavior of an LLM can be directly calibrated via the model's temperature setting — lower temperatures are known to lead to more consistent model output, and we test its impact on diversity.

The second set of methods (Study 3b) involves prompt engineering and the development of creative agents. LLM research has demonstrated that the specific wording, structure, and examples provided in prompts can dramatically improve LLM outputs (Zhou et al. 2023; De Freitas et al. 2025). One such technique, Chain-of-Thought (CoT) prompting, instructs the LLM to reason in multiple steps before concluding its answer, improving performance in many benchmarks (Wei et al. 2023).

Further, Osborn's (1953) foundational work on brainstorming emphasized the principle of "wild ideas" and deferring judgment to maximize creative output. Given that decades of brainstorming research have encouraged humans to create such bold ideas and thereby have provided them with strategies of how to explore the idea space, it is plausible that such an approach might also work in the case of AI. Specifically, we propose two creative agents that aim to restore the AI-induced diversity loss reported in Study 2. The first one instructs the LLM to expand on its current ideas by making them bolder and the second one combines pain points identified by humans with solutions generated by AI.

The third set of methods (Study 3c) takes a different approach by exploiting a key economic advantage of LLMs: their ability to generate ideas at near-zero marginal cost. Rather than improving diversity, we ask whether generating a much larger number of AI ideas can match the coverage of the human idea space.

### 4.1. Study 3a: Models, Parameters and Pooling

**Method:** We use the human-generated ideas from Study 1 as a benchmark. We also use our LLM-generated ideas from Study 1 (gpt-4-0314, 100 zero-shot and 100 few-shot ideas) and newly generated ideas (zero-shot) created by different models, including reasoning models. Compared to traditional LLMs, reasoning models first produce an internal step-by-step "thought process" before arriving at a final answer, a technique shown to improve performance across multiple benchmarks (OpenAI 2024). We generate new ideas using a broad array of frontier language models spanning multiple vendors and release dates, including both standard and reasoning variants (see Table 2 for details).

For each model, we used default hyperparameters. We first generate 2,000 ideas (20 batches of 100) and then randomly pick 100 ideas for comparison to control for underlying LLM output randomness.

We further test different temperature variations using GPT-4o, as it is one of the most widely used models. Temperature is a numerical value (typically between 0.7 and 1) that adjusts the likelihood of the model choosing less-probable tokens during text generation. In our experiments, we find that values greater than 1.0 often lead to unstable output, and even at 1.0, the output can become nonsensical on rare occasions. For our analysis, we use temperature settings of 0, 0.5, and 1.

To obtain a composite pool, we combine all ideas across the four leading vendor models (GPT-4 Turbo, Gemini Pro 2.0, DeepSeek R1 and Claude Sonnet 3.5) into a single idea pool of 4 models x 2,000 ideas per model = 8,000 ideas and draw a random balanced sample of 100 ideas (i.e., 25 ideas from each model). For each model, we calculate the diversity score as one minus the similarity score.

**Results:** Our results, summarized in Table 2, show that the tested LLMs are improving over time at creating diverse ideas. All but one model produced significantly more diverse results than the old GPT-4 model from March 2023, with o3-mini achieving the lowest diversity score out of all tested models. A careful inspection of its output revealed a curious fixation, with a disproportionately high number of ideas centered around desk organizers. The other reasoning model in our comparison, DeepSeek R1, however, placed near the top with very diverse results and clustered together closely with Gemini Pro 2.0 and GPT-4 Turbo. In addition, despite being a smaller model, GPT-3.5 Turbo ranked 8th in our comparison, ahead of the original (and older, see timeline in Figure S3) GPT-4. Notably, no single model came close to human diversity. However, the composite approach was as good as humans ($B$ = -0.010; 95% CI [-0.022, 0.002]; $t(198) = -1.69$, $p = .092$). See Table S18 for regression analysis results.

**Table 2.** Diversity Analysis (1 - Average Pairwise Cosine Similarity) Across Different Models (Higher Is Better)

| Rank | Model | Diversity [95% CI] | Release Date |
|---|---|---|---|
| 1 | Humans (Study 1) | 0.773 [0.764, 0.782] | |
| 2 | Composite pool (Random Subset) | 0.763 [0.755, 0.772] | |
| 3 | GPT-4 Turbo | 0.756 [0.747, 0.765] | April 2024 |
| 4 | Gemini Pro 2.0 | 0.752 [0.743, 0.761] | February 2025 |
| 5 | DeepSeek R1 | 0.752 [0.743, 0.761] | January 2025 |
| 6 | DeepSeek V3 | 0.700 [0.691, 0.708] | December 2024 |
| 7 | Claude Sonnet 3.5 | 0.682 [0.673, 0.691] | June 2024 |
| 8 | GPT-3.5 Turbo | 0.680 [0.671, 0.689] | June 2023 |
| 9 | GPT-4o | 0.679 [0.670, 0.688] | August 2024 |
| 10 | GPT-4 zero-shot (Study 1) | 0.585 [0.576, 0.594] | March 2023 |
| 11 | GPT-4 few-shot (Study 1) | 0.572 [0.563, 0.581] | March 2023 |
| 12 | o3-mini | 0.495 [0.486, 0.503] | January 2025 |

*Notes*. Higher diversity scores are better (lower similarity). Release dates refer to the public release of the specific model version tested. Exact model versions: GPT-4 Turbo (gpt-4-turbo-2024-04-09), Gemini Pro 2.0 (google-gemini-2-0-pro-exp-02-05),

DeepSeek R1 (reasoning variant of DeepSeek V3), Claude Sonnet 3.5 (claude-3-5-sonnet-20240620), GPT-3.5 Turbo (gpt-3.5-turbo-16k), GPT-4o (gpt-4o-2024-08-06), GPT-4 zero-shot and few-shot from Study 1 (gpt-4-0314), and o3-mini (o3-mini-2025-01-31). See Table S18 for more regression analysis results.

In addition, we found that altering the temperature parameter for GPT-4o has a significant effect on diversity, ranging from a diversity of 0.679 (temperature 1) to 0.618 (temperature 0.0, see Table S19).

**Discussion:** Study 3a provides early evidence that the diversity limitations of AI-generated ideas identified in Studies 2a and 2b may be partly addressed as models become more powerful. Interestingly, the strategy of creating a composite idea pool by mixing ideas across models works extremely well and almost reaches human-level diversity.

The substantial improvements achieved by models like DeepSeek R1, Gemini Pro 2.0, and GPT-4 Turbo represent practically significant advances that bring AI-generated idea diversity much closer to human levels, though still falling short of full parity. Figure S3 shows an overall trend in the ability of LLMs to create diverse sets of ideas in our setting. Overall, several later-released models generated more diverse pools than GPT-4-0314, although performance was heterogeneous and not monotonic in release date. However, the future evolution of this trend depends on how model vendors balance the desire for diversity with other dimensions of model performance.

Unsurprisingly, higher temperature values produce more creative output. However, at values above 1.0 the model output often becomes nonsensical and thereby increases in idea diversity come at the expense of idea quality. Beyond temperature, other architectural choices in how prompts are constructed may also affect idea generation in ways that merit further investigation (Brucks and Toubia 2025).

For innovation managers, our results indicate that experimentation is important to find the right model for the task. Perhaps most significantly, the success of the composite pooling approach represents a promising strategy for overcoming individual model limitations and reducing time spent evaluating models.

### 4.2. Study 3b: Prompting and Creative Agents

**Method:** We used GPT-4o (August 2024) as our reference model to test various interventions designed to increase idea diversity. Specifically, we investigated the following interventions:

- CoT: LLM first comes up with short idea drafts, then modifies them with the goal of increasing diversity before producing final product ideas (see E-Companion F for full prompt). We perform this in smaller steps of 10 ideas each (while maintaining previous context) as the model does not always follow all subtasks for larger sets of ideas at once.
- Entropy: We inject additional entropy into the generation process aiming to elicit more diverse generations. Specifically, we generate 100 ideas each across 20 sessions where we include (i) 20 randomly generated personas (e.g., a nursing student who works night shifts), (ii) a random functional design constraint (e.g., “the product must fold flat to fit in an envelope”) or (iii) a random sonnet from Shakespeare. See E-Companion G for full prompts and E-Companion H for a range of eccentric personas and constraints (e.g., from MBA students to Martians).

- Independent Generation: Previous experiments compared one LLM generating many ideas to many students generating one or a few ideas. Here, we generate one idea per independent LLM session.
- Agent 1 (Bold Ideas): LLM gets a random pool of 100 ideas and is tasked with altering each of the ideas by making it bolder. We achieve this by instructing the LLM “to make this idea bolder” and repeat this alteration of each of the 100 ideas two times. The final idea pool is the set of the altered ideas. Figure S6 (top) provides an example illustration.
- Agent 2 (Human-picked problem): LLM is presented with 100 student ideas and tasked with creating 10 product ideas that solve the same problem identified by the human but with a different solution. We randomly select one of these 10 ideas (see Figure S6 bottom).

**Results:** We found that diversity for GPT-4o can be substantially increased through our interventions (see Table 3). Chain-of-Thought prompting meaningfully increased idea diversity compared to the original GPT-4, though its improvement over GPT-4o was more modest, suggesting that explicit step-by-step reasoning helps models explore more diverse solution spaces. In addition, injecting heterogeneous personas and functional constraints significantly improved diversity. Adding sonnets into the prompt resulted in much smaller effects, suggesting that the type of introduced heterogeneity matters. Using independent LLM sessions that each generated just one idea led to a sharp decrease in diversity.

The two agent-based approaches revealed very contrasting outcomes. Against our expectations, the "Bold Ideas" agent, which iteratively refined ideas to make them bolder, greatly reduced diversity. Qualitative analysis revealed that many ideas converged on similar characteristics, particularly scents, suggesting that repeated refinement pushes LLMs toward a limited set of perceived novel features rather than true diversity. This pattern resembles local search becoming trapped on a suboptimal peak in a rugged solution landscape, where incremental refinement cannot escape to more distant, potentially superior regions (Wooten 2022). In contrast, humans working in tandem with an AI agent (human identifies the problem, AI generates the solution) achieved the second strongest performance out of all strategies, indicating that a human-directed search can be a viable strategy to retain diverse output. This supports the finding of higher diversity in human-directed search processes in Boussioux et al. (2024). However, instead of injecting heterogeneity by relying on human-generated pain points, other mechanisms can be used, in particular Table 3 suggests that obtaining the heterogeneity from random personas is an equally effective strategy. However, humans working without AI support still produce the most diverse ideas. See Table S21 for regression analysis results.

**Table 3.** Diversity Analysis (1 - Average Pairwise Cosine Similarity) Across Different Strategies

| Rank | Model | Diversity [95% CI] |
|---|---|---|
| 1 | Humans (Study 1) | 0.773 [0.764, 0.782] |
| 2 | Random Personas | 0.757 [0.748, 0.766] |
| 3 | Agent 2 (human-picked problem) | 0.739 [0.730, 0.748] |
| 4 | Random Constraints | 0.735 [0.726, 0.744] |
| 5 | GPT-4o Chain-of-Thought | 0.698 [0.689, 0.707] |
| 6 | Random Sonnets | 0.686 [0.677, 0.695] |
| 7 | GPT-4o zero-shot | 0.679 [0.670, 0.688] |
| 8 | GPT-4 zero-shot, Study 1 | 0.585 [0.576, 0.594] |
| 9 | Agent 1 (refine ideas 2 times) | 0.575 [0.566, 0.584] |
| 10 | GPT-4o Independent Sessions | 0.452 [0.442, 0.461] |

*Notes*. Higher diversity scores are better (lower similarity). See Table S21 for more regression analysis results.

Further, prompting using more eccentric personas and constraints did not monotonically increase diversity (see E-Companion H for details). In a 2D space, more eccentric prompts pushed ideas farther from human ideas (see Figure 5). However, they still clustered heavily, now in regions underexplored by humans.

**Figure 5.** Pushing AI Toward Eccentricity Spreads Ideas Outward but Clusters Remain

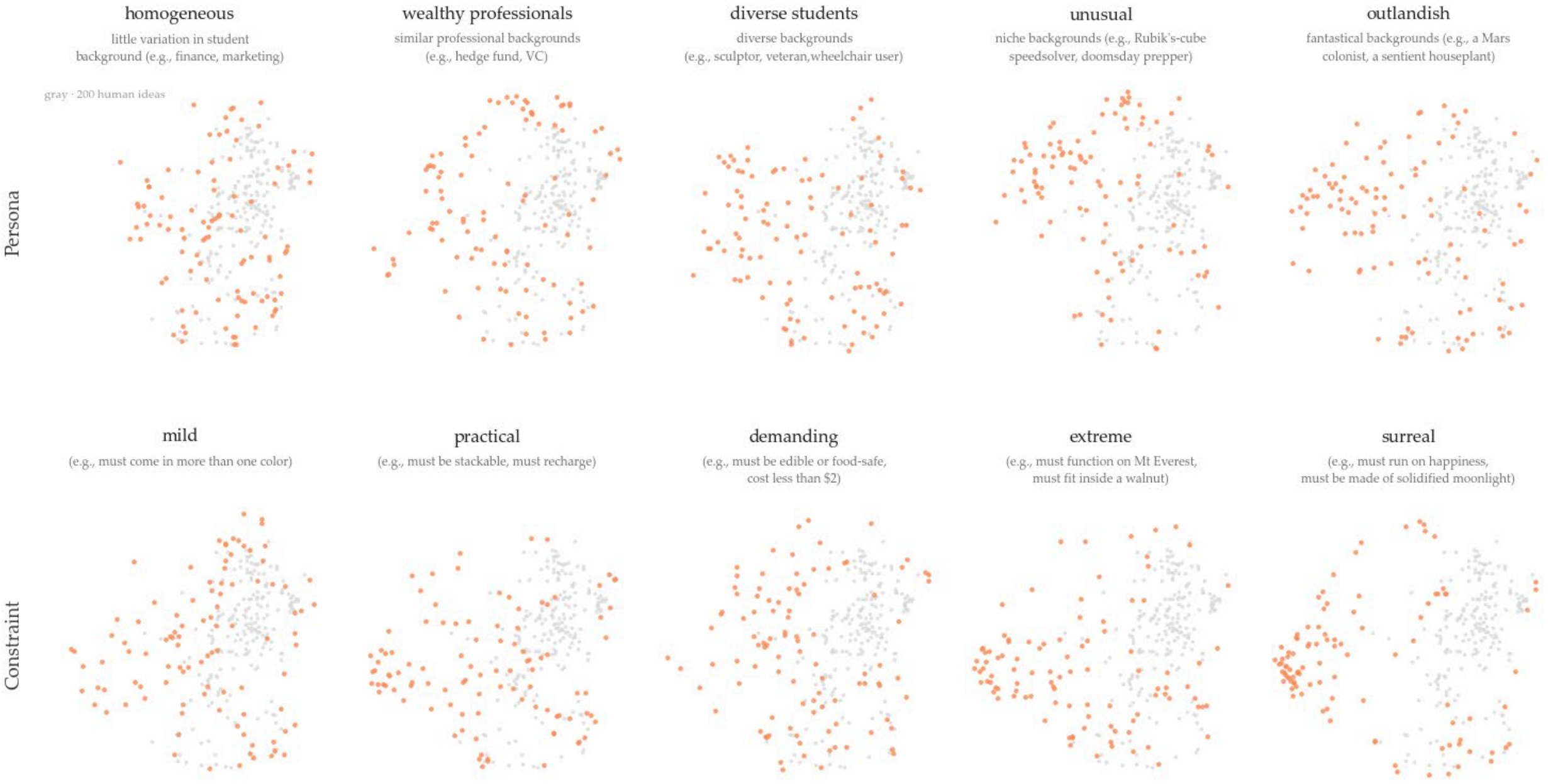

*Notes.* Figure shows 200 human ideas (gray) and 100 LLM-generated ideas each (orange) for different personas and constraints (see E-Companion H for details). The middle set is the same as reported in Table 3. The ideas are projected from their original USE-based embedding (512-dimensions) down to shared two-dimensional UMAP space; display only.

**Discussion:** Similar to Study 3a, we find that the diversity limitations of AI-generated ideas represent significant challenges but are not insurmountable barriers. Through targeted interventions, particularly human-guided problem exploration, the diversity of AI-generated ideas can be improved substantially.

Our finding that repeated refinement led to convergence on common "novel" features like scents suggests that AI systems may have implicit biases about what constitutes novelty or boldness and warrants caution for unsupervised idea refinement. At the same time, injecting task-relevant heterogeneity proved as effective as human-guided problem exploration, suggesting that diversity can also be recovered by forcing the model to adopt varied starting points rather than relying solely on human direction. Importantly, not all forms of entropy are equally useful, indicating that the heterogeneity may have to connect to the ideation task. However, our results provide practical guidance for diverse ideas by involving humans in the ideation process to identify pain points. When humans identify specific problems or pain points and direct AI systems to explore multiple solutions within those spaces, the result is both a more diverse and a more targeted exploration of the solution landscape. Picking a human identified pain point for idea generation is akin to providing rapid feedback in an innovation contest, a practice that Mihm and Schlapp (2019) show to improve performance. This finding supports a collaborative model of AI-assisted creativity where humans and AI systems play complementary roles: humans excel at problem identification, space definition, and diversity evaluation, while AI systems excel at rapid solution generation within defined parameters. Such a division of labor leverages the strengths of both human and artificial intelligence while compensating for their respective limitations.

## 4.3. Study 3c: Scaling the Number of Ideas

The analyses in Studies 2 and 3a-b compare equal-sized pools of human and AI ideas, but in practice a manager can generate thousands of AI ideas at negligible cost. We therefore ask whether simply generating more AI ideas can close the coverage gap between AI-generated and human-generated ideas.

**Method:** Diversity, our primary metric so far, captures the average pairwise distance within an idea pool. Coverage, by contrast, captures how well the AI idea pool spans the space occupied by ideas—conceptually, the area covered in the embedding space illustrated in Figure 3. If we keep adding ideas into that space, we should eventually cover the same regions human ideas occupy.

We operationalize coverage using nearest-neighbor cosine distances: for each of the 200 human ideas, we compute its cosine distance to the closest idea in an LLM pool of increasing size. Lower mean and maximum distances indicate better coverage of the human idea space through the LLM-generated ideas. Using GPT-4o ideas from the previous studies, we created cumulative random subsets of N = 100, 200, 300, 500, 750, and 1,000 ideas, measuring coverage at each step.

**Results**: Increasing the number of AI ideas steadily improves coverage of the human idea space (see Figure S7). The mean nearest-neighbor distance from human ideas to the closest LLM idea drops from 0.546 at N = 100 to 0.478 at N = 1,000. This decline shows no sign of plateauing within the range tested, suggesting that further scaling would continue to close the gap. The maximum distance—the hardest-to-match human idea—declines more slowly (from 0.762 to 0.711), indicating that some human ideas occupy regions of the idea space that may remain difficult for the LLM to reach even at scale.

**Discussion**: Study 3c highlights that the practical implications of the diversity gap depend on how managers use AI in their ideation process. When the goal is to explore the full landscape of possible ideas—as in an innovation tournament—generating a large, inexpensive pool of AI ideas and then curating the most promising ones is a viable strategy. Our results suggest that coverage improves steadily with scale and even at 1,000 has not yet plateaued. At the same time, the slower decline in maximum distance suggests that certain regions of the human idea space may be systematically harder for LLMs to reach in our setting—an interesting phenomenon that deserves future investigation.

## 5. Conclusion

Our research provides a detailed empirical assessment of how large language models, such as GPT-4, perform in the task of new product idea generation. We find that ideas generated by an LLM score higher in average purchase intent than those generated by humans. This result holds across different prompting methods and is particularly pronounced in the few-shot prompting condition. Notably, AI-generated ideas are seven times more likely to fall into the top decile of ideas based on purchase intent. These findings suggest that generative AI can serve as a high-throughput and high-yield idea-generation engine in early-stage innovation processes. As Studies 1b and 1c demonstrate, this performance reflects superior idea quality, not better pitching skills or regurgitating existing ideas.

We also document a critical limitation of AI-powered brainstorming in our setting: a loss of idea diversity. Compared to human-generated idea sets, AI-generated ideas exhibit greater semantic similarity. Using text mining and pairwise similarity metrics, we show that the tested LLMs tend to converge on a narrower region of the solution space. The diversity loss suggests that these idea sets may provide a less varied portfolio of conceptual bets, potentially leaving some regions of the idea space underexplored and lowering the likelihood of identifying conceptually distinct or radical ideas. This limits the exploration of the idea space and reduces the probability of identifying conceptually distinct or radical innovations. We further generalize this result by reanalyzing four prior studies using LLMs for idea generation, confirming that reduced diversity is a robust empirical pattern.

In Studies 3a and 3b, we propose several techniques that innovation managers can use to counter this diversity loss. First, we find that, with few exceptions, idea diversity suffers less when more recent

LLMs are deployed. Second, choosing the right model and parameter settings does make a difference. In case of doubt, pooling ideas across models from different vendors seems like an easy-to-implement strategy. Third, we show that prompt engineering techniques also increase diversity. Similarly, injecting structured heterogeneity into the generation process, such as random personas or functional design constraints, significantly boosts diversity, though not all forms of randomness are equally effective. Finally, and somewhat more technically, building "creative agents" that actively explore distinct regions of the idea space can restore diversity to near-human levels in our setting. Of particular interest are agents that leverage human insight for unaddressed needs and then use LLMs to find creative solutions to those needs. This suggests a new division of labor in AI-augmented innovation: LLMs excel at generating high-quality ideas but are most effective when humans direct them to specific open needs.

Beyond improving diversity, we show in Study 3c that managers can exploit the near-zero marginal cost of AI idea generation by simply producing a much larger pool. Coverage of the human idea space improves steadily as the number of AI ideas increases, with no sign of diminishing returns at 1,000 ideas, though some regions of the human idea space remain systematically harder to reach. This provides innovation managers with a complementary, low-effort strategy: rather than optimizing any single model or prompt, generate at scale and select.

We acknowledge several limitations of our work that warrant attention in future research. First, our main analysis is based on a single tournament seeking new product ideas for the college market. Future work should aim to increase the generalizability of our findings by replicating them across different innovation tournaments and across a variety of innovation challenges (such as other new product development settings and alternative innovation tasks). Second, we did not explicitly examine the interaction between humans and AI during the process of idea generation. Although in Study 3b we speculated about the potential benefits of having AI generate solutions for needs identified by humans, our human participants did not directly experience this workflow. It would be valuable to explore whether exposure to high-quality AI solutions might lead human innovators to target a different set of unmet needs. Third, we represented ideas solely through text. In practice, skilled human product developers make extensive use of visual representations, ranging from simple sketches to sophisticated digital prototypes. Would such visual representations also enhance the creativity of AI-powered or fully automated development? Finally, we note that innovation involves not only generating ideas but also selecting the best ones and advancing them toward implementation. After all, the value of generating a great idea is lost if a mediocre one is ultimately chosen for development. Future research could investigate the extent to which AI is capable of evaluating and selecting both AI- and human-generated ideas.

## Acknowledgments

The authors gratefully thank the reviewers of POM. We thank the Mack Institute for Innovation Management for generous research support.

## References

Ayers, J. W., A. Poliak, M. Dredze, E. C. Leas, Z. Zhu, J. B. Kelley, D. J. Faix, A. M. Goodman, C. A. Longhurst, M. Hogarth, D. M. Smith. 2023. Comparing Physician and Artificial Intelligence Chatbot Responses to Patient Questions Posted to a Public Social Media Forum. *JAMA Internal Medicine* 183(6), 589–596.

Bai, H., J. G. Voelkel, S. Muldowney, J. C. Eichstaedt, R. Willer. 2025. LLM-generated messages can persuade humans on policy issues. *Nature Communications* 16(1), 6037.

Baruah, J., P. B. Paulus. 2011. Category assignment and relatedness in the group ideation process. *Journal of Experimental Social Psychology* 47(6), 1070–1077.

Berger, J., G. Packard, R. Boghrati, M. Hsu, A. Humphreys, A. Luangrath, S. Moore, G. Nave, C. Olivola, M. Rocklage. 2022. Marketing insights from text analysis. *Marketing Letters* 33(3), 365–377.

Boudreau, K. J., N. Lacetera, K. R. Lakhani. 2011. Incentives and Problem Uncertainty in Innovation Contests: An Empirical Analysis. *Management Science* 57(5), 843–863.

Boussioux, L., J. N. Lane, M. Zhang, V. Jacimovic, K. R. Lakhani. 2024. The Crowdless Future? Generative AI and Creative Problem-Solving. *Organization Science* 35(5), 1589–1607.

Bronnec, F. L., A. Verine, B. Negrevergne, Y. Chevaleyre, A. Allauzen. 2024, June 4. Exploring Precision and Recall to assess the quality and diversity of LLMs. arXiv.

Brucks, M. S., S.-C. Huang. 2020. Does practice make perfect? The contrasting effects of repeated practice on creativity. *Journal of the Association for Consumer Research* 5(3), 291–301.

Brucks, M., O. Toubia. 2025. Prompt architecture induces methodological artifacts in large language models. *PLOS ONE* 20(4), e0319159.

Bubeck, S., V. Chandrasekaran, R. Eldan, J. Gehrke, E. Horvitz, E. Kamar, P. Lee, Y. T. Lee, Y. Li, S. Lundberg, H. Nori, H. Palangi, M. T. Ribeiro, Y. Zhang. 2023, April 13. Sparks of Artificial General Intelligence: Early experiments with GPT-4. arXiv.

Chen, S., O. Urminsky, G. Zhang, R. Walatka, K. Fernandez, A. Low, J. Bogard, C. R. Fox. 2026, March 2. Estimating the threat of AI-agent responding across online survey platforms. PsyArXiv.

Costello, T. H., G. Pennycook, D. G. Rand. 2024. Durably reducing conspiracy beliefs through dialogues with AI. *Science* 385(6714), eadq1814.

Cox, S. R., Y. Wang, A. Abdul, C. Von Der Weth, B. Y. Lim. 2021. Directed Diversity: Leveraging Language Embedding Distances for Collective Creativity in Crowd Ideation. *Proceedings of the 2021 CHI Conference on Human Factors in Computing Systems*. Yokohama, Japan, ACM, 1–35.

Dahan, E., H. Mendelson. 2001. An Extreme-Value Model of Concept Testing. *Management Science* 47(1), 102–116.

De Freitas, J., G. Nave, S. Puntoni. 2025. Ideation with Generative AI—in Consumer Research and Beyond. *Journal of Consumer Research* 52(1), 18–31.

Dell'Acqua, F., E. McFowland, E. Mollick, H. Lifshitz, K. C. Kellogg, S. Rajendran, L. Krayer, F. Candelon, K. R. Lakhani. 2026. Navigating the Jagged Technological Frontier: Field Experimental Evidence of the Effects of Artificial Intelligence on Knowledge Worker Productivity and Quality. *Organization Science* 37(2), 403–423.

Doshi, A. R., O. P. Hauser. 2024. Generative AI enhances individual creativity but reduces the collective diversity of novel content. *Science Advances* 10(28), eadn5290.

Gielnik, M. M., M. Frese, J. M. Graf, A. Kampschulte. 2012. Creativity in the opportunity identification process and the moderating effect of diversity of information. *Journal of Business Venturing* 27(5), 559–576.

Girotra, K., C. Terwiesch, K. T. Ulrich. 2010. Idea Generation and the Quality of the Best Idea. *Management Science* 56(4), 591–605.

Gordon, A., D. Rothschild, F. M. Affonso, J. Sulik, D. Hauser, K. Pepin, S. Jones. 2026, April 1. AI Agent Prevalence and Data Quality Across Multiple Online Sample Providers. PsyArXiv.

Haase, J., P. H. P. Hanel. 2023. Artificial muses: Generative artificial intelligence chatbots have risen to human-level creativity. *Journal of Creativity* 33(3), 100066.

Haase, J., P. H. P. Hanel, S. Pokutta. 2025, April 10. Has the Creativity of Large-Language Models peaked? An analysis of inter- and intra-LLM variability. arXiv.

Herbold, S., A. Hautli-Janisz, U. Heuer, Z. Kikteva, A. Trautsch. 2023. A large-scale comparison of human-written versus ChatGPT-generated essays. *Scientific Reports* 13(1), 18617.

Hubert, K. F., K. N. Awa, D. L. Zabelina. 2024. The current state of artificial intelligence generative language models is more creative than humans on divergent thinking tasks. *Scientific Reports* 14(1), 3440.

Jamieson, L. F., F. M. Bass. 1989. Adjusting Stated Intention Measures to Predict Trial Purchase of New Products: A Comparison of Models and Methods. *Journal of Marketing Research* 26(3), 336–345.

Johnson, D. R., A. S. Cuthbert, M. E. Tynan. 2021. The neglect of idea diversity in creative idea generation and evaluation. *Psychology of Aesthetics, Creativity, and the Arts* 15(1), 125–135.

Joosten, J., V. Bilgram, A. Hahn, D. Totzek. 2024. Comparing the Ideation Quality of Humans With Generative Artificial Intelligence. *IEEE Engineering Management Review* 52(2), 153–164.

Kalvapalle, S. G., N. Phillips, J. Cornelissen. 2024. Entrepreneurial Pitching: A Critical Review and Integrative Framework. *Academy of Management Annals* 18(2), 550–599.

Katz, D. M., M. J. Bommarito II, S. Gao, P. Arredondo. 2024. GPT-4 Passes the Bar Exam. *Philosophical Transactions of the Royal Society A* 382(2254).

Kirk, R., I. Mediratta, C. Nalmpantis, J. Luketina, E. Hambro, E. Grefenstette, R. Raileanu. 2024, February 19. Understanding the Effects of RLHF on LLM Generalisation and Diversity. arXiv.

Koenker, R., K. F. Hallock. 2001. Quantile Regression. *The Journal of Economic Perspectives* 15(4), 143–156.

Koivisto, M., S. Grassini. 2023. Best humans still outperform artificial intelligence in a creative divergent thinking task. *Scientific Reports* 13(1), 13601.

Kornish, L. J., J. Hutchison-Krupat. 2017. Research on Idea Generation and Selection: Implications for Management

of Technology. *Production and Operations Management* 26(4), 633–651.

Kornish, L. J., K. T. Ulrich. 2011. Opportunity Spaces in Innovation: Empirical Analysis of Large Samples of Ideas. *Management Science* 57(1), 107–128.

Kornish, L. J., K. T. Ulrich. 2014. The Importance of the Raw Idea in Innovation: Testing the Sow's Ear Hypothesis. *Journal of Marketing Research* 51(1), 14–26.

Lee, B. C., J. (Jae) Chung. 2024. An empirical investigation of the impact of ChatGPT on creativity. *Nature Human Behaviour* 8(10), 1906–1914.

Lee, B. C., J. (Jae) Chung. 2025. Reply to: ChatGPT decreases idea diversity in brainstorming. *Nature Human Behaviour* 9(6), 1110–1112.

Lifshitz-Assaf, H., S. Lebovitz, L. Zalmanson. 2021. Minimal and Adaptive Coordination: How Hackathons' Projects Accelerate Innovation without Killing it. *Academy of Management Journal* 64(3), 684–715.

Matz, S. C., J. D. Teeny, S. S. Vaid, H. Peters, G. M. Harari, M. Cerf. 2024. The potential of generative AI for personalized persuasion at scale. *Scientific Reports* 14(1), 4692.

McCoy, R. T., P. Smolensky, T. Linzen, J. Gao, A. Celikyilmaz. 2023. How Much Do Language Models Copy From Their Training Data? Evaluating Linguistic Novelty in Text Generation Using RAVEN. *Transactions of the Association for Computational Linguistics* 11: 652–670.

Meincke, L., G. Nave, C. Terwiesch. 2025. ChatGPT decreases idea diversity in brainstorming. *Nature Human Behaviour 9(6), 1107–1109.*

Mihm, J., J. Schlapp. 2019. Sourcing Innovation: On Feedback in Contests. *Management Science* 65(2), 559–576.

Miller, A. I. 2019. *The Artist in the Machine: The World of AI-Powered Creativity*. Cambridge, MA, MIT Press.

Newell, A., J. C. Shaw, H. A. Simon. 1959. Report on a General Problem-Solving Program. *Proceedings of the International Conference on Information Processing*. Paris, UNESCO, 256–264.

Nguyen, T. 2024, November 5. Understanding Transformers via N-gram Statistics. arXiv.

OpenAI. 2023, March 14. GPT-4. GPT-4 is OpenAI's most advanced system, producing safer and more useful responses. Available at https://openai.com/index/gpt-4/ (accessed date July 1, 2025).

OpenAI. 2024, December 21. OpenAI o1 System Card. arXiv.

Osborn, A. F. 1953. *Applied Imagination: Principles and Procedures of Creative Thinking*. New York, Charles Scribner's Sons.

Peng, H., H. S. Qiu, H. B. Fosse, B. Uzzi. 2024. Promotional language and the adoption of innovative ideas in science. *Proceedings of the National Academy of Sciences* 121(25).

Poetz, M. K., M. Schreier. 2012. The Value of Crowdsourcing: Can Users Really Compete with Professionals in Generating New Product Ideas? *Journal of Product Innovation Management* 29(2), 245–256.

Redden, J. P. 2025. Test for the Best: Using ChatGPT to Create Effective Ad Taglines. *Journal of Current Issues & Research in Advertising* 46(2), 123–140.

Rietzschel, E. F., B. A. Nijstad, W. Stroebe. 2007. Relative accessibility of domain knowledge and creativity: The effects of knowledge activation on the quantity and originality of generated ideas. *Journal of Experimental Social Psychology* 43(6), 933–946.

Salvi, F., M. Horta Ribeiro, R. Gallotti, R. West. 2025. On the conversational persuasiveness of GPT-4. *Nature Human Behaviour* 9(8), 1645–1653.

Shibayama, S., D. Yin, K. Matsumoto. 2021. Measuring novelty in science with word embedding. *PLOS ONE* 16(7), e0254034.

Si, H., S. Kavadias, C. Loch. 2022. Managing innovation portfolios: From project selection to portfolio design. *Production and Operations Management* 31(12), 4572–4588.

Siangliulue, P., J. Chan, S. P. Dow, K. Z. Gajos. 2016. IdeaHound: improving large-scale collaborative ideation with crowd-powered real-time semantic modeling. *Proceedings of the 29th Annual Symposium on User Interface Software and Technology*. 609–624.

Simon, H. A. 1967. Understanding Creativity. J. C. Gowan, G. D. Demos, & E. P. Torrance, eds. *Creativity: Its Educational Implications*. New York, John Wiley & Sons.

Sommer, S. C., C. H. Loch. 2004. Selectionism and Learning in Projects with Complexity and Unforeseeable Uncertainty. *Management Science* 50(10), 1334–1347.

Stevenson, C., I. Smal, M. Baas, R. Grasman, H. van der Maas. 2022, June 10. Putting GPT-3's Creativity to the (Alternative Uses) Test. arXiv.org. Retrieved June 14, 2025, https://arxiv.org/abs/2206.08932v1.

Terwiesch, C., K. T. Ulrich. 2009. *Innovation Tournaments: Creating and Selecting Exceptional Opportunities*. Boston, MA, Harvard Business Press.

Toubia, O. 2006. Idea Generation, Creativity, and Incentives. *Marketing Science* 25(5), 411–425.

Turing, A. M. 1950. Computing Machinery and Intelligence. *Mind* 59(236), 433–460.

Ulrich, K., S. Eppinger. 2007. *Product Design and Development*. McGraw-Hill Education.

Van der Stigchel, S., C. Strauch, B. de Zwart, L. Van Maanen. 2026. Will online behavioral research follow the fate of online survey research? *Proceedings of the National Academy of Sciences* 123(8), e2535585123.

Wei, J., X. Wang, D. Schuurmans, M. Bosma, B. Ichter, F. Xia, E. Chi, Q. Le, D. Zhou. 2023, January 10. Chain-of-Thought Prompting Elicits Reasoning in Large Language Models. arXiv.

Weitzman, M. L. 1979. Optimal Search for the Best Alternative. *Econometrica* 47(3), 641–654.

Westwood, S. J. 2025. The potential existential threat of large language models to online survey research. *Proceedings of the National Academy of Sciences* 122(47), e2518075122.

Westwood, S. J., S. Frederick. 2026. Reply to Van der Stigchel et al.: Empirical evidence that AI survey contamination is real and substantial. *Proceedings of the National Academy of Sciences* 123(8), e2537420123.

Wooten, J. O. 2022. Leaps in innovation and the Bannister effect in contests. *Production and Operations Management* 31(6), 2646–2663.

Wooten, J. O., K. T. Ulrich. 2017. Idea Generation and the Role of Feedback: Evidence from Field Experiments with Innovation Tournaments. *Production and Operations Management* 26(1), 80–99.

Zhou, D., N. Schärli, L. Hou, J. Wei, N. Scales, X. Wang, D. Schuurmans, C. Cui, O. Bousquet, Q. Le, E. Chi. 2023, April 16. Least-to-Most Prompting Enables Complex Reasoning in Large Language Models. arXiv

# E-Companion for AI and Its Impact on Creativity and Diversity: An Empirical Study of LLM-Generated Product Ideas

## A. Prompting

We prompt OpenAI's GPT-4 (more specifically, gpt-4-0314) with the same prompt we gave the students. No LLM yet acts entirely autonomously. Rather, they are tools used by humans to complete tasks. For this study, we aim for minimal prompt engineering, thus representing a novice user scenario. However, we acknowledge that many strategies could potentially improve LLM performance. For instance, Mihm and Schlapp (2019) show that providing feedback during ideation contests can further improve performance of human innovators and we expect this to hold for LLMs as well.

For our first LLM-generated idea pool (zero-shot) we use the system prompt to provide contextual information and subsequent user prompts to ask for ideas, ten at a time. The user prompt includes the additional request that the descriptions be 40-80 words, like the student sample.

*System Prompt*

*"You are a creative entrepreneur looking to generate new product ideas. The product will target college students in the United States. It should be a physical good, not a service or software. I'd like a product that could be sold at a retail price of less than about USD 50. The ideas are just ideas. The product need not yet exist, nor may it necessarily be clearly feasible. Number all ideas and give them a name. The name and idea are separated by a colon."*

*User Prompt*

*"Please generate ten ideas as ten separate paragraphs. The idea should be expressed as a paragraph of 40-80 words."*

The model used is gpt-4-0314 with the "temperature" parameter at 0.7 to retain randomness and thus greater creativity. The temperature parameter controls the randomness of the output, with lower values leading to more deterministic output and higher values increasing variability. At the time of the experiment, the suggested default value for temperature was 0.7 to strike a balance between coherence and creativity, without possibly sampling highly unlikely tokens (i.e., semantic chunks used for representational efficiency) that lead to undesirable responses.

An obstacle to using GPT-4 for generating hundreds of ideas is its finite memory, typically limited to the number of tokens the underlying LLM can consider in generating its responses. Once the number of tokens in a session exceeds the model's limit, the LLM has no memory of the first ideas generated, and subsequent ideas can become increasingly redundant. The number of tokens in the version of GPT-4 we had access to was about 8,000, roughly 7,000 words or approximately 80 ideas (some tokens are used for the system and user prompt and idea titles).

To generate more than 80 ideas resulting from the limited context window, we asked GPT-4 to "compress" the previously generated ideas into shorter summaries. These summaries were then provided to the model before generating the next batch of ideas, ensuring that the model knows the previously generated ideas while remaining within the context limits. We used the below summarization prompt, followed by the original system prompt and generated summaries, and finally, a user prompt that explicitly asks for different ideas.

*Summarization Prompt*

*"Aggressively compress the following ideas so that their original meaning remains but they are much shorter. You can use tags or keywords. : <Ideas generated so far> "*

*System Prompt*

*<Original System Prompt> + "Previously you generated the following ideas and should not repeat them: <Summaries> "*

*User Prompt*

*<Original User Prompt> + "Make sure they are different from the previous ideas."*

For our second pool of LLM-generated ideas (few-shot), we provide the LLM with examples (few-shot learning) of high-quality ideas generated by students. In particular, we appended our prompts to provide the LLM with six highly rated ideas from a separate student set that completed the same exercise and informed GPT-4 that these ideas had been well-received by students in our class. We used six examples due to context window limitations at the time of the experiment.

*Good Ideas Prompt*

*<Original System Prompt>* + "Here are some well received ideas for inspiration: *<Good Ideas>*"

## B. Idea Rephrasing Prompt

We include six examples from our previous LLM-generated ideas to illustrate the desired style:

"StudySerenity Noise-Canceling Earplugs: High-quality, reusable earplugs designed to help college students block out distractions and focus on their studies. These earplugs are made from soft, hypoallergenic materials that provide a comfortable fit and effective noise reduction. With a compact carrying case, StudySerenity Earplugs can be easily stored and transported, making them an essential tool for students who need peace and quiet in shared living spaces or noisy environments.

CollegeSurvival Multitool: A versatile and compact multitool designed to assist college students with a range of everyday tasks and challenges. The CollegeSurvival Multitool includes essential tools such as a bottle opener, scissors, screwdriver, and pen, all housed in a durable casing. This practical and portable device provides students with a reliable solution for various situations, from fixing a loose screw on a desk to opening a package from home.

CollegeLife Multifunctional Chair: This innovative, space-saving chair for college students is designed to be both functional and stylish. With its adjustable backrest and foldable design, the chair can be easily transformed into a desk, a lounge chair, or even a bed for those late-night study sessions. Built-in storage compartments provide ample space for books, stationery, and other essentials, while its lightweight and portable design allows for easy transportation and storage in tight dorm rooms.

Nightstand Nook: A compact, multifunctional nightstand designed for dorm rooms with limited space. The Nightstand Nook features a built-in wireless charging pad for smartphones, a USB charging hub, a slide-out cup holder, and a storage compartment for essentials like glasses, earbuds, and medication. Its lightweight, modular design allows it to be easily attached to a bed frame or wall and blends seamlessly with most decor styles.

Ultimate Study Snack Box: Keep energy levels high during long study sessions with the Ultimate Study Snack Box. This curated collection of healthy, delicious snacks is perfect for college students who want quick, nutritious options to fuel their brains. The snack box includes a variety of options, such as protein bars, nuts, dried fruits, and more, all packaged in a compact, reusable container. This makes it easy for students to grab a snack and stay focused, whether they're in their dorm room or at the library.

The Study Bumper: A portable, inflatable study booth designed to create a private, noise-reducing space for college students in shared living arrangements. The Study Bumper is easy to set up, lightweight, and can be quickly deflated and stored when not in use. It features a built-in desk with compartments for stationery and electronics, as well as a comfortable, ergonomic seat. This innovative study solution helps students maintain focus and concentration during their study sessions, even in busy dormitories or apartments.

Consider the examples above and rephrase the product description below in a similar style and length. Do not add any details unless they are clearly specified in the original description."

## C. Idea Descriptions

**Idea A: BookBuddy**

*A versatile, portable book holder designed to make reading and studying more comfortable and efficient for college students. The BookBuddy features adjustable arms that securely hold textbooks, tablets, and laptops at the perfect angle for ergonomic viewing. Its lightweight, foldable design makes it easy to carry in a backpack, while its non-slip base ensures stability on various surfaces. The BookBuddy is an essential tool for academic success in any setting.*

**Idea B: Adjustable Laptop Riser**

*Many college students spend long hours working on their laptops, leading to poor posture and potential health problems. The Adjustable Laptop Riser is a portable, ergonomic solution designed to elevate laptops to a more comfortable viewing angle, promoting better posture and reducing eye strain. With its durable construction and non-slip surface, this riser can support a variety of laptop sizes and weights. Its compact, foldable design makes it easy to transport and store, making it an essential accessory for any college student.*

**Idea C: Hydration Harness**

*In 2023, water bottles are becoming increasingly prevalent in the classroom as gen z is waking up to the importance of hydration. As students, our backpacks are all equipped with side pouches to hold them. But for some of us, particularly gym goers and athletes, these pouches are just too small to securely hold our large water bottles. Holding my water can be annoying and cumbersome, as in my busy life I often want my hands free. Hydration Harness is an adjustable strap that can fit any size water bottle and seamlessly and stylishly attach to your existing backpack.*

## D. USE Details

To represent each idea as a numerical vector, we used the Universal Sentence Encoder (USE), a pre-trained model that maps variable-length text to fixed-length 512-dimensional embedding vectors. USE is trained on a variety of natural language tasks and data sources, producing embeddings that capture semantic similarity: texts with related meanings are mapped to nearby points in the embedding space, while semantically distinct texts are mapped further apart. We chose USE because it has been widely adopted in creativity and ideation research (e.g., Dell'Acqua et al. 2026) and provides a standardized, reproducible representation that does not require task-specific fine-tuning. After embedding each idea, we computed pairwise cosine similarity between embedding vectors as our primary measure of idea similarity (following Cox et al. 2021 and Dell'Acqua et al. 2026).

## E. Rephrasing Details

A potential concern with using embedding-based similarity to measure idea diversity is that the embedding model may conflate stylistic variation with topical variation. Human-generated ideas in our sample exhibit considerable stylistic heterogeneity, use first-person pronouns, rhetorical questions, exclamations, and colloquial language. GPT-4-generated ideas, by contrast, are stylistically more uniform (written in third-person declarative prose at consistent lengths). If the embedding model encodes these stylistic differences as semantic dissimilarity, the observed diversity gap between human and AI-generated ideas could be partially or wholly an artifact of writing style rather than a reflection of genuine topical diversity.

To address this concern, we applied a symmetric standardization procedure to all 400 ideas (200 human, 100 GPT-4 zero-shot, 100 GPT-4 few-shot) using claude-sonnet-4-5. For each idea, we applied the prompt below alongside two independently generated samples that were not part of the 200 ideas. If the output violated the word count constraint (within 5 words), the request was retried up to two times:

"You are a text standardization tool. You rephrase product idea descriptions
to match a consistent, neutral, specification-oriented voice while preserving content and length.

Your output should read as if GPT-4 had originally generated the idea. Study these two
real GPT-4 examples carefully and match their tone, structure, and vocabulary:

EXAMPLE 1:
Title: Eco-Pen Set

Description: A stylish set of biodegradable and refillable pens made from recycled materials. Each set includes five pens in different exciting colors and designs, perfect for environmentally-conscious students. The ink is made from soy-based materials, ensuring smooth writing and minimal impact on the environment. Package these in a reusable bamboo case that doubles as a storage box.

EXAMPLE 2:
Title: Portable Snack Peeler
Description: A compact, handheld tool designed to peel fruits and vegetables easily, right in the dorm. It's made with a silicone grip and stainless steel blade for safe and efficient use. Perfect for students trying to eat healthier snacks, it supports simple food prep in limited kitchen spaces.

Key patterns to replicate:
- Opens with "A/The [adjective], [adjective] [noun]..." or "A/The [adjective] set/tool/device..."
- Uses concise, direct sentences that describe what the product is and what it does.
- Mentions materials and design specifics naturally (e.g., "silicone grip", "stainless steel blade", "soy-based materials").
- Third person only. No "I", "you", "we", "my", "your". "It's" for "it is" is acceptable.
- Declarative statements only. No questions, no exclamations.
- Practical tone — not overly formal, not casual. Reads like a product listing.
- Specification-oriented: describe what the product IS and DOES.

LENGTH RULES (CRITICAL):
- Your description MUST be within +/- 5 words of the original word count. We are aiming at around 60 words per idea, so you may slightly expand or condense as needed, but do NOT deviate more than 5 words in either direction.
- Do NOT pad short ideas with filler.
- Do NOT invent features or details that aren't in the original.
- Do NOT compress long ideas. Keep all the original's content.

TITLE RULES:
- Compound brand-name style (e.g., "QuickClean Mini Vacuum", "FreshAlert Fridge Timer").
- Keep the title short (2-4 words).

Return ONLY the rephrased title and description in this exact format:

Title: [rephrased title]
Description: [rephrased description]”

All original and standardized idea descriptions were embedded using Google USE. For each idea within a group, we computed its mean pairwise cosine similarity to all other ideas in the same group. The resulting per-idea similarity score was then averaged across all ideas in the group to yield the within-group similarity metric. To assess how much semantic content was preserved by the standardization, we computed a fidelity score for each idea: the cosine similarity between its original and standardized embeddings. Higher fidelity indicates that the standardization changed the style without altering the underlying meaning.

The diversity gap between human and GPT-4 ideas persisted after standardization. Initially, the zero-shot gap was large and highly significant ($B = 0.193$, 95% CI [0.182, 0.204], $t(298) = 33.89$, $p < .001$). After standardization, this effect remained highly significant ($B = 0.154$, 95% CI [0.141, 0.168], $t(298) = 21.82$, $p < .001$), retaining 80% of the original gap. Adding word count as a covariate produced only a marginal further reduction ($B = 0.149$, 95% CI [0.136, 0.161], $t(297) = 23.27$, $p < .001$). Results for the few-shot comparison were similar: the original condition effect ($B = 0.207$, 95% CI [0.196, 0.219], $t(298) = 35.98$, $p < .001$) remained highly significant after standardization ($B = 0.186$, 95% CI [0.172, 0.201], $t(298) = 25.69$, $p < .001$), retaining 90% of the original gap. Controlling for word count produced a modest further reduction ($B = 0.177$, 95% CI [0.164, 0.190], $t(297) = 26.68$, $p < .001$). The 10-20% reduction may be attributable to the removal of human stylistic diversity (rhetorical questions, personal anecdotes, colloquial phrasing) that the embedding model encoded as additional dissimilarity. The large residual gap thus reflects genuine differences in the topical content of ideas generated by humans versus GPT-4.

## F. Chain-of-Thought Prompt

“You are a creative entrepreneur looking to generate new product ideas. The product will target college students in the United States. It should be a physical good, not a service or software. I'd like a product that could be sold at a retail price of less than about USD 50. The ideas are just ideas. The product need not yet exist, nor may it necessarily be clearly feasible. Number all ideas and give them a name. The name and idea are separated by a colon.

First generate a list of 10 ideas.
Second, go through the list and determine whether the ideas are different and bold, modify the ideas as needed to make them more bold and different from each other.

Next, give the ideas a name and combine it with a product description. The name and idea are separated by a colon and followed by a description. The idea should be expressed as a paragraph of 40-80 words. Do this step by step!"

## G. Entropy Prompting Approaches

The diversity gap documented in E-Companion E persists even after stylistic standardization, suggesting that GPT-4 ideas converge on a relatively narrow region of the space of student-targeted physical products (water bottles, study aids, dorm organizers, etc.). A natural follow-up question is whether this is a fundamental property of LLM ideation or an artifact of how we prompt the model. If the latter, then injecting additional entropy at generation time — varying who the LLM is, what design constraints it must respect, or what stimulus it has just attended to — should pull samples toward different regions of idea space and shrink the gap. We tested three variants of entropy injection. Each samples a stochastic element from a fixed pool and embeds it in the prompt. We hold all other parameters constant: model gpt-4o, temperature 1.0, and the original system/user prompt structure described in E-Companion A. For each variant we run 20 independent sessions, generate 100 ideas per session, and randomly sample 100 ideas from the resulting pool of 2,000 to keep the comparison size-matched to the 100 GPT-4 zero-shot ideas in Study 3. We have three conditions, with the base system/user prompt being identical to E-Companion A. The three variants modify only the prefix:

1. Persona: Before being asked to generate ideas, the model is instructed to adopt a randomly drawn persona (e.g., "a nursing student who works night shifts," "an aerospace engineering student in a rocketry club"). The persona pool contains 100 student personas spanning majors, extracurriculars, life circumstances, and backgrounds. Each session draws a unique persona without replacement. The persona is intended to shift the model's representation of the target user and the experiences it draws on when ideating. Prompt: "You are {persona} looking to generate new product ideas. The product will target college students in the United States…"
2. Functional constraint: The model receives an additional functional requirement that any proposed product must satisfy (e.g., "The product must fold flat to fit in an envelope," "The product must work without electricity or batteries," "The product must serve at least two completely different functions"). The constraint pool contains 20 such requirements, each used once per session. Constraints force the model away from its modal solutions by ruling out the easiest answers. Prompt appended after the system and before the user prompt: Additional constraint: {constraint}.
3. Shakespeare sonnet: Each session is preceded by a randomly drawn Shakespearean sonnet from Project Gutenberg's eBook edition of Shakespeare's Sonnets (Shakespeare 1997). This is a content-irrelevant entropy injection: the sonnet has no semantic connection to product design, but exposes the model to fresh tokens that may steer subsequent generation through the residual stream. The sonnet is appended before the system prompt with no instruction other than the original ideation task.

## H. Persona and Constraint Variations

We tested a five-level eccentricity gradient for persona and constraints following our existing method (100 ideas drawn from a sample of 2000).

For personas, the levels ranged from homogeneous (near-identical first-year business undergraduates, e.g., "a first-year accounting major on the CPA track"), through mostly homogeneous wealthy professionals (e.g., a venture capitalist, hedge fund managers), the college-student set with very diverse background as a default (e.g., a pre-med biology major, a music-composition major who plays in a jazz band, a veteran in a wheelchair), real-but-unusual personas (e.g., a falconer, a competitive yo-yo champion), to outlandish fictional or non-human personas (e.g., a Mars colonist, a time traveler, a sentient cloud). In short, as far as personas are concerned, we measure the degree of eccentricity on the following scale: homogenous -> wealthy professionals -> diverse students -> unusual -> outlandish

For constraints, the gradient ran from mild near-defaults ("must come in more than one color"), through practical demands ("must be stackable," "must recharge over USB-C"), functional but demanding constraints ("must be edible or food-safe," "must cost less than $2 to manufacture"), extreme but physically possible demands ("must function at the summit of Mount Everest," "must fit inside a walnut shell"), to surreal, physically impossible constraints ("must run on happiness," "must be made of solidified moonlight"). In short, as far as constraints are concerned, we measure the degree of eccentricity on the following scale: mild -> practical -> demanding -> extreme -> surreal.

The results show that the strategy and its execution cannot be separated and eccentricity showed no clean monotonic relationship with diversity. All five levels of each strategy remained significantly less diverse than the human baseline (all $p < .001$, except the extreme-constraint tier at $p < .01$). For personas, the homogeneous set was significantly the least diverse (vs. every other tier, all $p < .001$), but pushing toward more eccentric personas did not increase diversity further: the diverse student set was significantly more diverse than the most eccentric one ($p < .001$) and numerically—though not significantly—more diverse than the second most eccentric ($p = .09$). For constraints, diversity peaked neither at the demanding set nor at the most eccentric level but at the second most eccentric level (extreme-yet-physically possible), which was significantly the most diverse condition—more diverse than the demanding set and every other level (all $p < .001$)—and the closest to the human baseline, though still significantly above it ($p < .01$); pushing one step further into surreal, impossible constraints significantly reversed the gain (second most vs. most eccentric, $p < .001$), returning diversity to a level statistically indistinguishable from the demanding set ($p = .51$). All p-values are uncorrected.

**Table H1.** Persona Eccentricity — Similarity Analysis (Average Pairwise Cosine Similarity) Across Eccentricity Tiers (Lower Is More Diverse)

| **Model** | **Coefficient** | **95% CI** | ***p*-value** |
|---|---|---|---|
| Intercept (Human ideas, Study 1) | 0.221 | [0.215, 0.227] | < .001 |
| Homogeneous (rank 1) | 0.088 | [0.077, 0.098] | < .001 |
| Wealthy Professionals (rank 2) | 0.048 | [0.037, 0.058] | < .001 |
| Diverse Students (Study 3b, rank 3) | 0.029 | [0.018, 0.039] | < .001 |
| Unusual (rank 4) | 0.037 | [0.027, 0.048] | < .001 |
| Outlandish (rank 5) | 0.055 | [0.044, 0.065] | < .001 |

*Notes*. $N = 700$. $F(5, 694) = 62.48$. OLS estimates using pairwise cosine-similarity (method 1). Baseline category for condition is the human student set (Study 1); all estimates therefore represent relative mean differences. Positive coefficients represent greater similarity.

**Table H2.** Constraint Eccentricity — Similarity Analysis (Average Pairwise Cosine Similarity) Across Eccentricity Tiers (Lower Is More Diverse)

| **Model** | **Coefficient** | **95% CI** | ***p*-value** |
|---|---|---|---|
| Intercept (Human ideas, Study 1) | 0.221 | [0.214, 0.227] | < .001 |
| Mild (rank 1) | 0.053 | [0.042, 0.064] | < .001 |
| Practical (rank 2) | 0.082 | [0.071, 0.093] | < .001 |
| Demanding (Study 3b, rank 3) | 0.044 | [0.033, 0.055] | < .001 |
| Extreme (rank 4) | 0.018 | [0.007, 0.029] | .001 |
| Surreal (rank 5) | 0.040 | [0.030, 0.051] | < .001 |

*Notes*. $N = 700$. $F(5, 694) = 52.58$. OLS estimates using pairwise cosine-similarity (method 1). Baseline category for condition is the human student set (Study 1); all estimates therefore represent relative mean differences. Positive coefficients represent greater similarity.

# I. Supplemental Figures and Tables

This section contains supplemental figures and tables.

## I.1. Supplemental Figures

**Figure S1.** Estimated Difference in Idea Quality Ratings between AI-generated Ideas and Human-generated ideas (baseline), for Different Percentiles

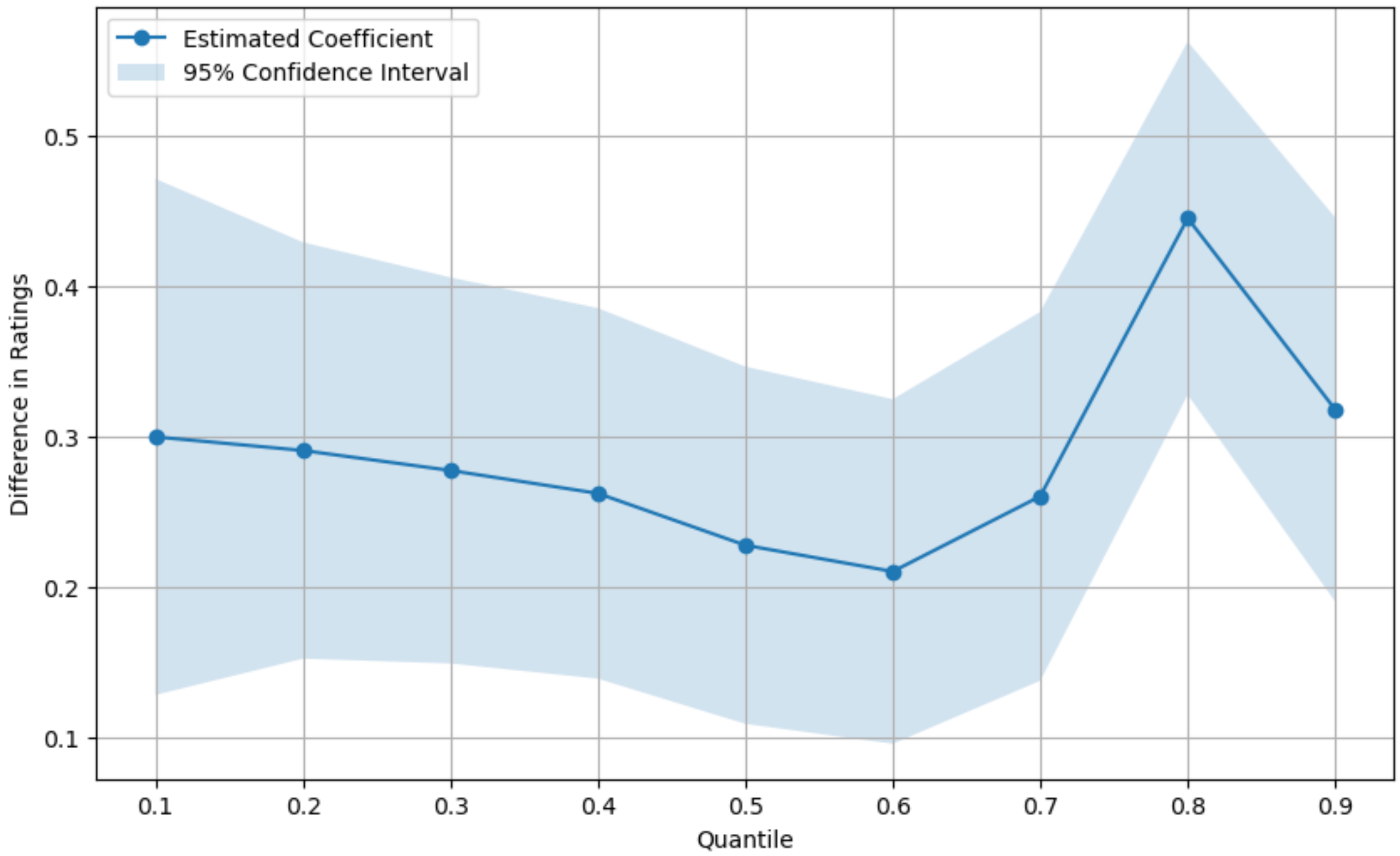

*Notes*. Positive values indicate a higher rating for GPT-generated ideas than for student ideas.

**Figure S2.** Estimated Difference in Idea Quality Ratings between AI-rephrased Ideas and Human-generated ideas (baseline), for Different Percentiles

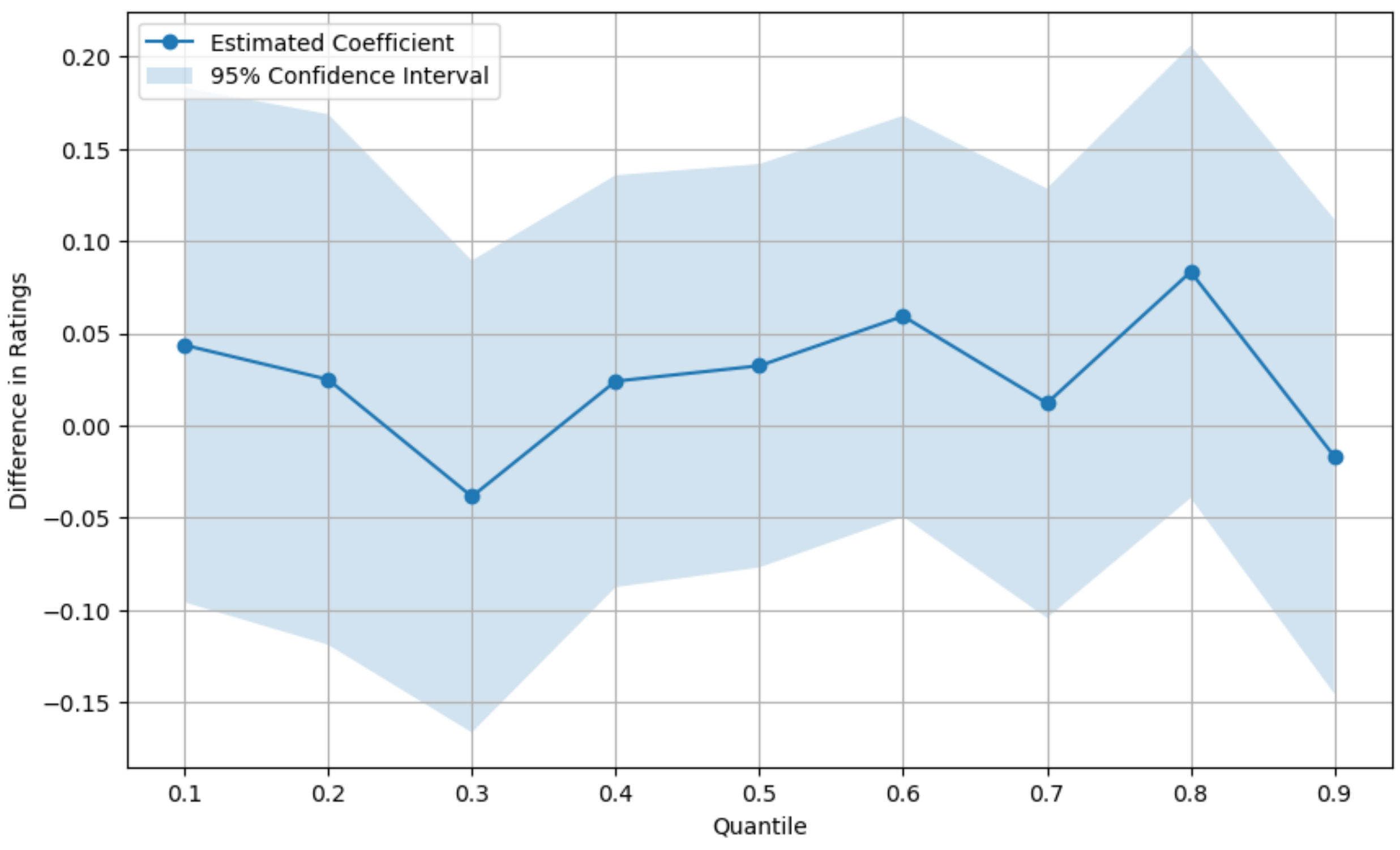

*Notes*. Positive values indicate a higher rating for GPT-rephrased ideas than for student ideas.

**Figure S3.** Models and Release Dates

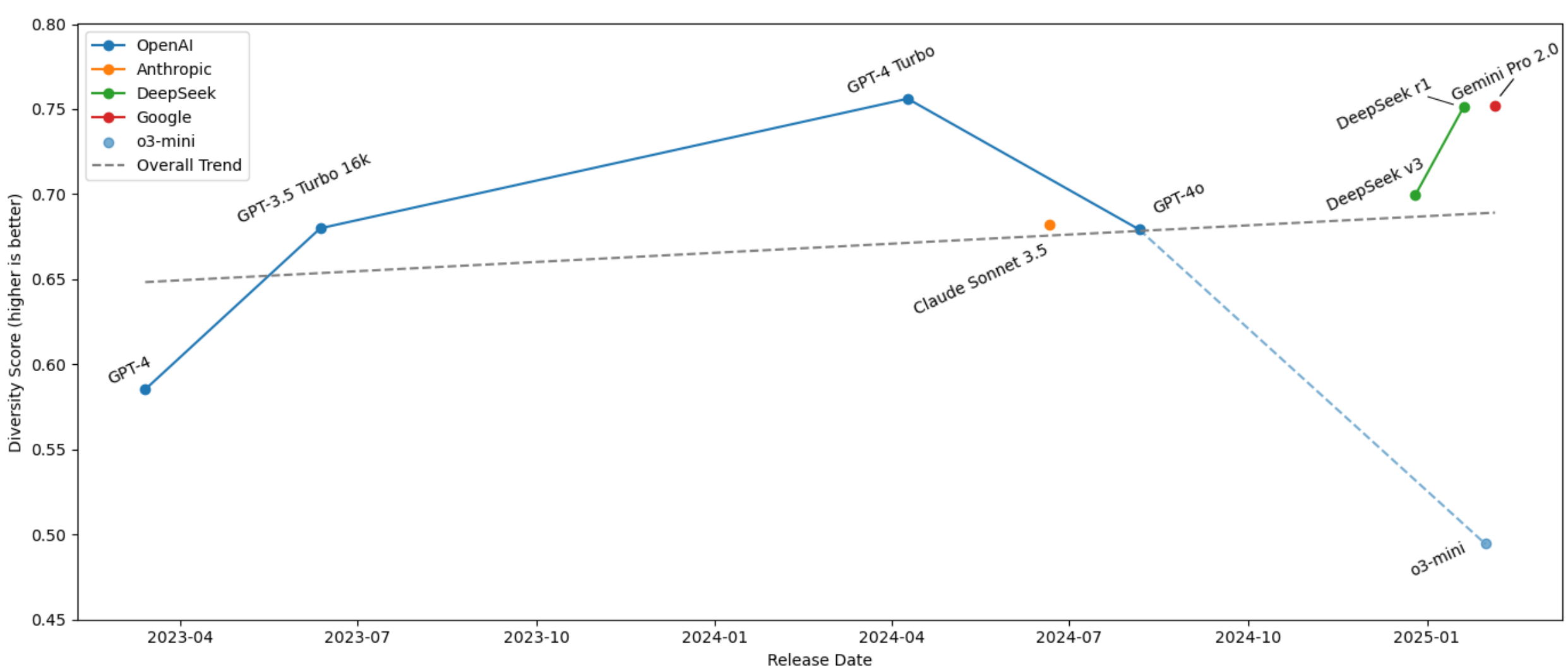

*Notes*. Two data points do not fit the overall trend of newer models leading to improved diversity. OpenAI may have sacrificed some creativity in their development of o3-mini and GPT-4o. Both models provided major advancements in the ability of the LLM to deal with other data than text. These advances suggest that creativity might not be a primary training objective in all models and so understanding the strengths and weaknesses of each model is important. In addition, our analysis is limited to random samples of 100 ideas from larger LLM pools and the variation in diversity scores may reflect not only genuine capability differences but also differential sensitivity to prompt architecture, which has been shown to affect models inconsistently (Brucks and Toubia 2025).

**Figure S4.** Pooled Median Split of Purchase Intent and Novelty by Source: Human vs AI Ideas (Joint Distribution by Quadrant)

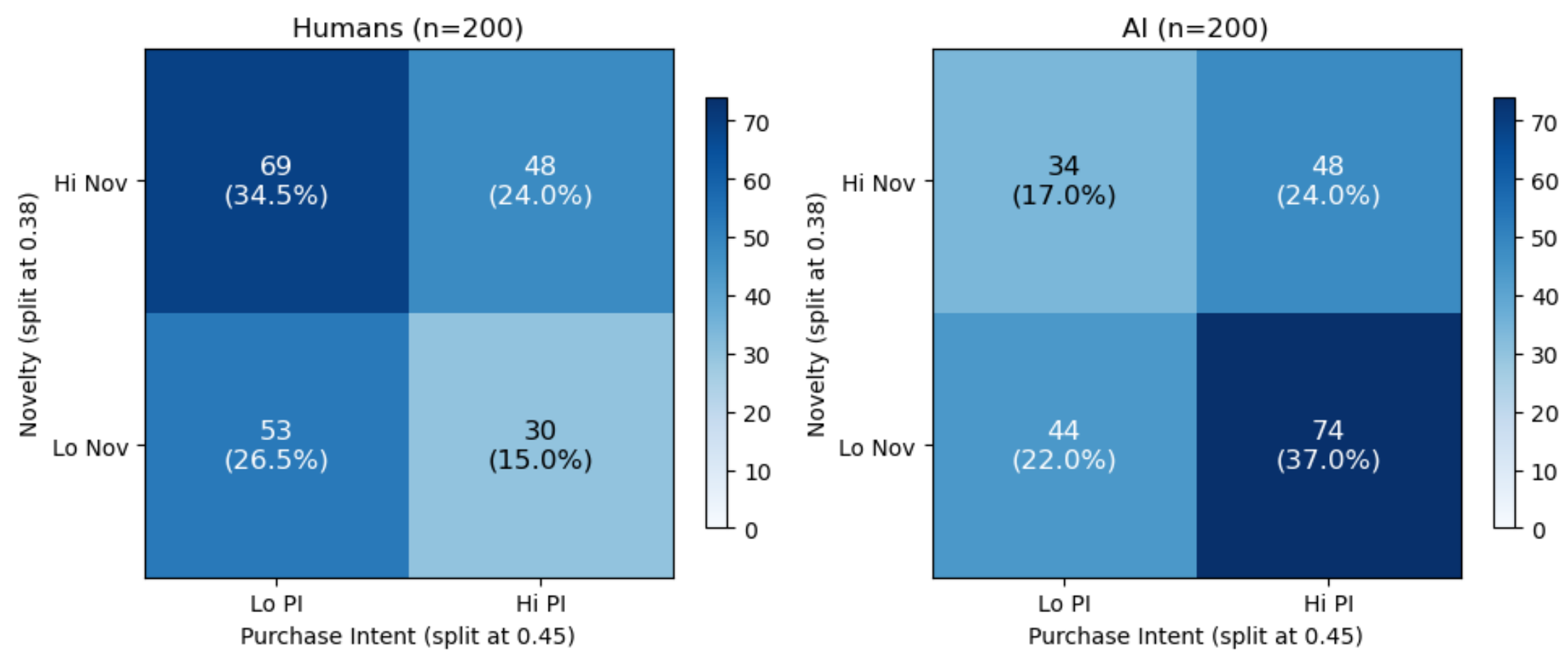

*Notes*. Humans and AI produced exactly the same number of high-purchase-intent/high-novelty ideas (48 in each group). The main difference was not in this "best of both worlds" quadrant. Instead, AI ideas were much more common in the high-purchase-intent/low-novelty quadrant, 74 AI ideas versus 30 human ideas, while human ideas were much more common in the low-purchase-intent/high-novelty quadrant, 69 human ideas versus 34 AI ideas.

**Figure S5.** Shift-function Analysis of Idea Quality (Left) and Novelty (Right) Across Human- and AI-generated ideas

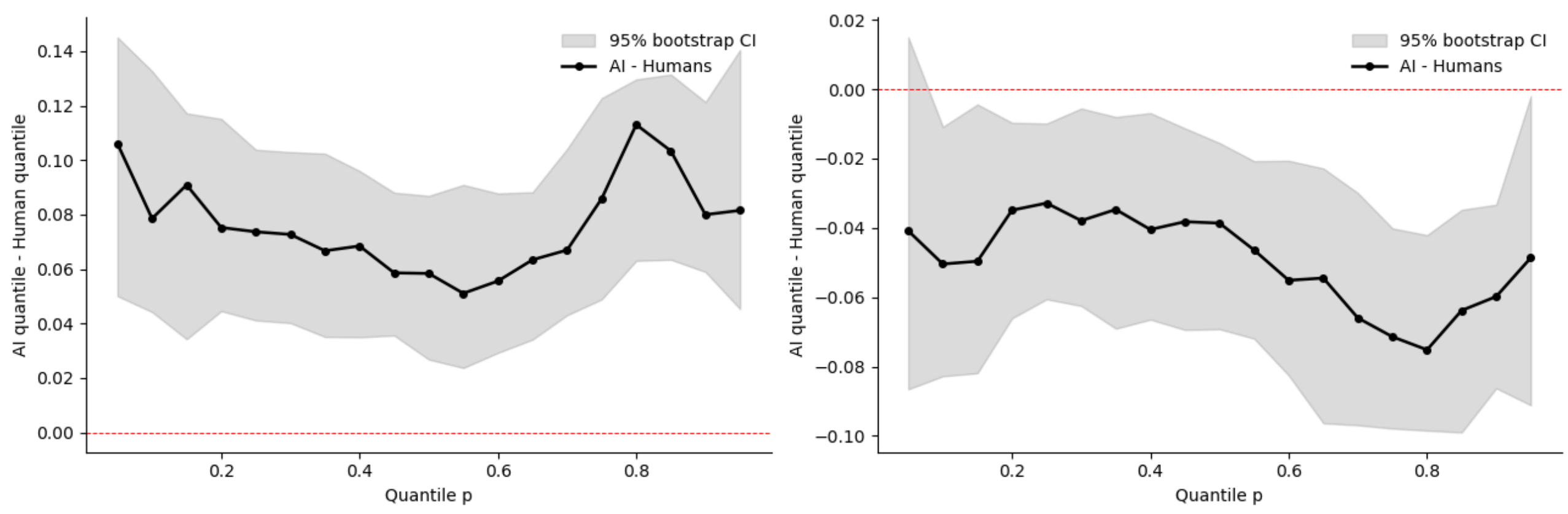

*Notes*. Purchase intent (left) and novelty (right) of human- and AI-generated ideas in the college setting across quantiles. Black lines indicate the difference between AI and human quality scores (left) and AI and human novelty scores (right) respectively. The red line indicates no difference. AI ideas were higher than human ideas on purchase intent across the entire distribution. The AI-human difference in purchase intent was positive at the 5th percentile (0.106), the median (0.058), the 75th percentile (0.086), and the 90th percentile (0.080), with bootstrapped 95% confidence intervals excluding zero throughout. The AI advantage appears closer to a broad upward shift in the purchase-intent distribution instead of specifically avoiding bad or generating more top-end ideas. The novelty distribution moved in the opposite direction. AI ideas were lower than human ideas on novelty across most quantiles, including the 25th percentile (-0.033), median (-0.039), 75th percentile (-0.071), and 90th percentile (-0.060).

**Figure S6.** Agentic Strategies

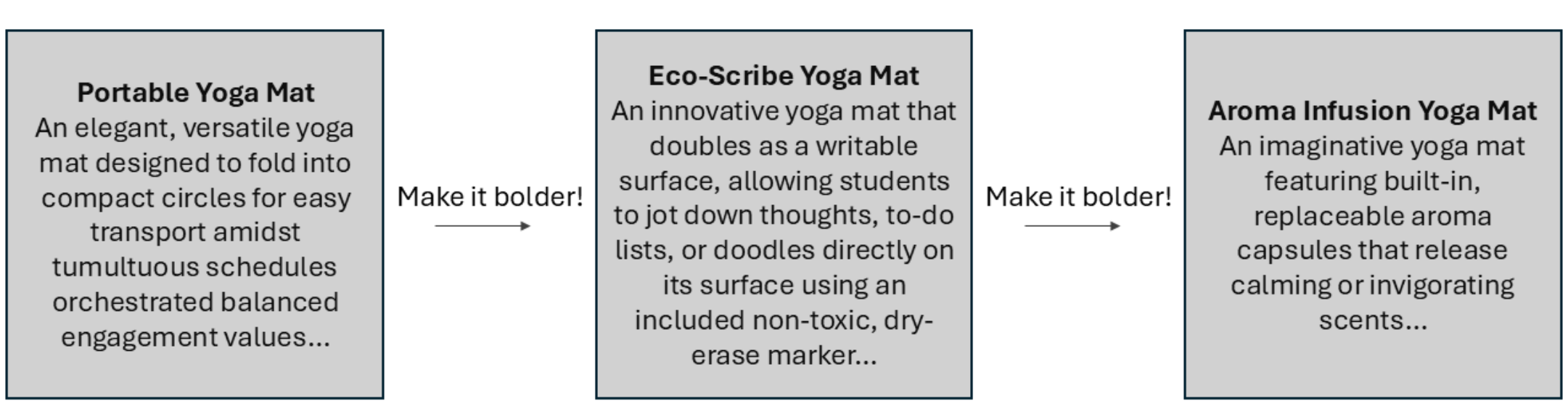

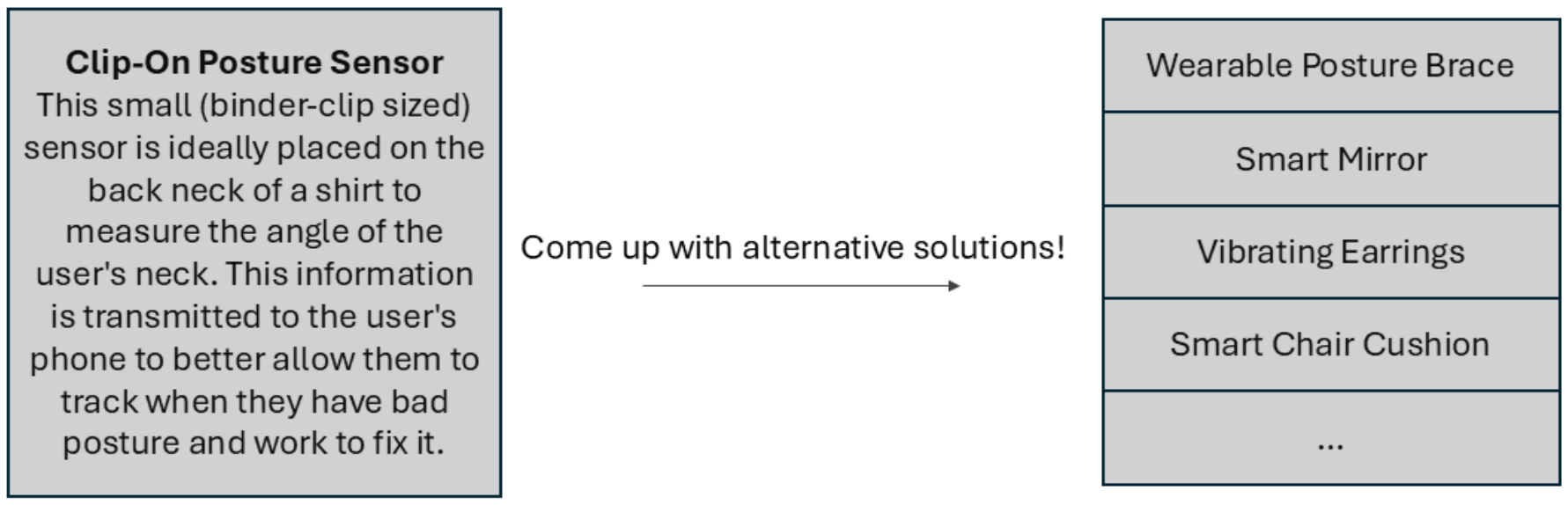

*Notes.* Agent 1 ideas are shortened for brevity. Agent 2 ideas only include titles for brevity.

**Figure S7.** Coverage of the Human Idea Space as the Number of LLM-Generated Ideas Increases

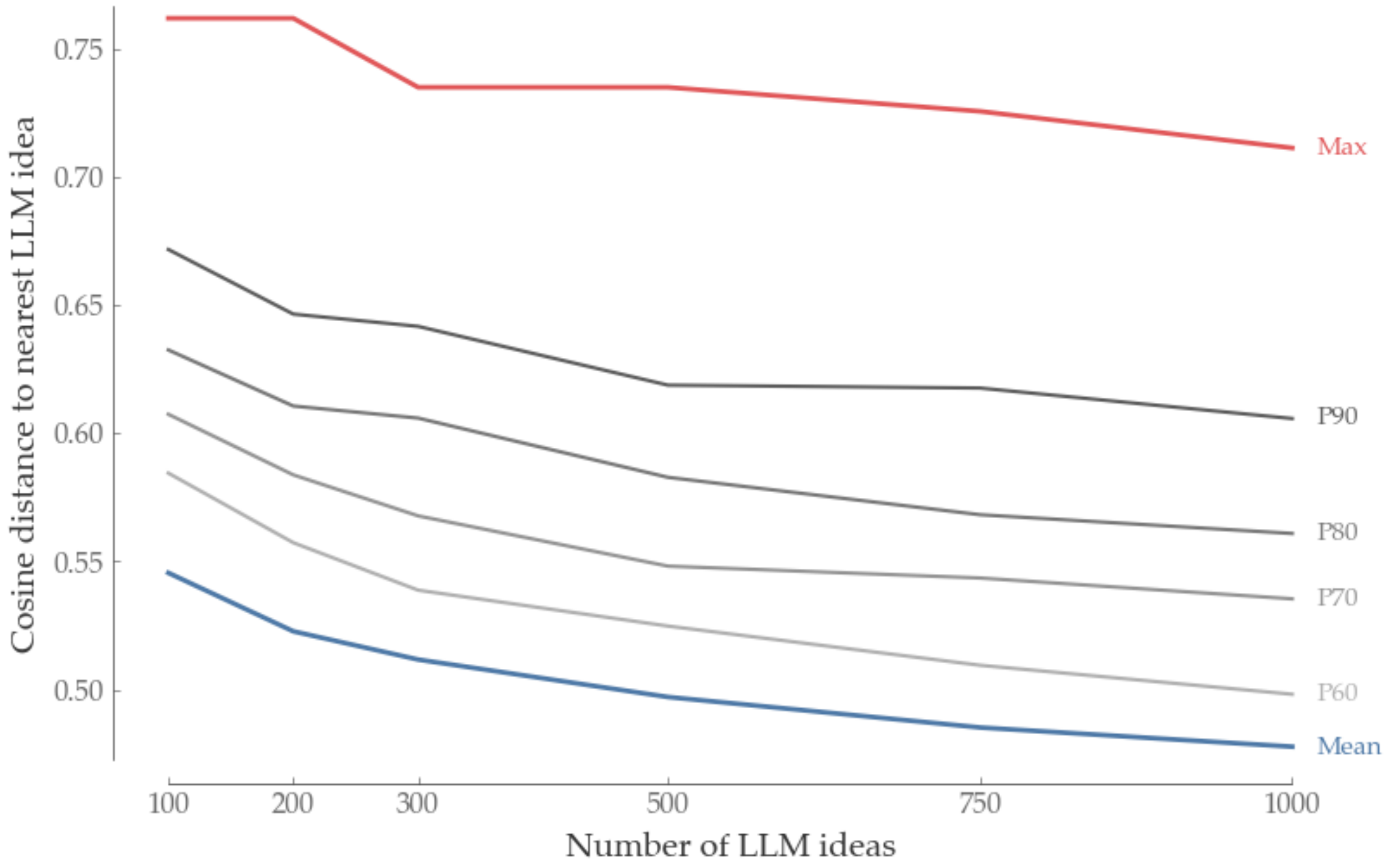

*Notes*. For each of the 200 human ideas, we compute the cosine distance to its nearest neighbor in an LLM pool (GPT-4o) of increasing size (N = 100 to 1,000). Lines show summary statistics of this nearest-neighbor distance distribution: mean (blue), percentiles (P60-P90, gray), and maximum (red). Lower values indicate better coverage of the human idea space.

## I.2. Supplemental Tables

**Table S1.** Overview of Studies, Research Aims, Contributions and Prior Work

| Study | Aim | Contribution | Prior work |
| --- | --- | --- | --- |
| 1a | What is the impact of using AI-powered brainstorming on the quality of the best idea? | Show that AI-powered brainstorming generates ideas that are of slightly higher average quality and dramatically more likely to be among the best ideas. | Haase and Hanel (2023) find LLMs reach human-level in divergent thinking; Hubert et al. (2024) show AI more creative than humans in CT/DAT tasks; Koivisto and Grassini (2023) find AI exceeds average but not best human ideas; Lee and Chung (2024) show ChatGPT users generate more creative ideas; Boussioux et al. (2024) find AI ideas have higher strategic viability; Joosten et al. (2024) show AI ideas score higher in novelty/benefit |
| 1b | How much of the observed advantage in Study 1a can be explained by superior AI pitching skills? | Show that the superior quality reported in 1a is not a result of the LLM's pitching skills. | Bai et al. (2025), Herbold et al. (2023), Matz et al. (2024), Salvi et al. (2025) show LLMs outperform humans in persuasion; Kalvapalle et al. (2024) review importance of pitching skills in entrepreneurship |
| 1c | How much of the observed advantage in Study 1a can be explained by the AI proposing existing (less novel) ideas? | Show that when prompted to generate new ideas, the LLMs are not simply regurgitating existing ideas. | Boussioux et al. (2024) find AI decreased novelty; Joosten et al. (2024) report increased novelty; McCoy et al. (2023) and Nguyen (2024) show LLMs can reproduce training data verbatim; Shibayama et al. (2021) provide novelty measurement framework |
| 2a | What is the impact of using AI-powered brainstorming on the diversity of the ideas generated in our study? | Show that AI-powered brainstorming generates sets of ideas that are less diverse. | Doshi and Hauser (2024) find AI-assisted creative writing less diverse; Meincke et al. (2025) confirm diversity loss analyzing Lee and Chung (2024) data; Cox et al. (2021) and Dell'Acqua et al. (2026) provide diversity measurement methodologies |
| 2b | What is the impact of using AI-powered brainstorming on the diversity of the ideas generated in prior studies? | Confirm that GenAI diversity loss is not unique to this study but generalizes across tasks and diversity metrics | Analyzes data from Boussioux et al. (2024), Hubert et al. (2024), and Stevenson et al. (2022) using multiple diversity metrics from Cox et al. (2021), Siangliulue et al. (2016), and Doshi and Hauser (2024) |
| 3a | By how much can the diversity of AI-generated idea sets be increased by vendor choice and model configuration? | Explore variations across LLM models and model configurations.<br><br>Show that cross-vendor pooling nearly restores human-level diversity; | Haase et al. (2025) find no significant creativity improvements in newer models using DAT assessment; OpenAI (2023) claims GPT-4 is "more creative than ever before" |
| 3b | By how much can the diversity of AI-generated idea sets be increased by prompting strategies and the use of creative agents? | Show that prompting and using creative agents using the LLM's API can help to restore diversity | Osborn (1953) emphasizes "wild ideas" in brainstorming; Wei et al. (2023) introduce Chain-of-Thought prompting; Zhou et al. (2023) show prompt engineering improves outputs; Boussioux et al. (2024) find higher diversity in human-directed search; De Freitas et al. (2025) suggest prompt-engineering |
| 3c | Can scaling the number of AI-generated ideas close the coverage gap with human idea sets? | Show that generating larger pools of AI ideas steadily improves coverage of the human idea space, though some regions remain systematically harder for LLMs to reach | Kornish and Ulrich (2011) estimate opportunity spaces contain thousands of unique ideas; Terwiesch and Ulrich (2009) discuss parallel search and innovation tournaments; Girotra et al. (2010) and Wooten and Ulrich (2017) emphasize the importance of exploring the full opportunity landscape |

**Table S2.** Estimated Difference in Purchase Intent Between Humans (baseline) and GPT-4 Zero-Shot and Few-Shot

| | **Purchase Intent Rating** | | |
|---|---|---|---|
| *Predictors* | *Estimates* | *CI* | *p* |
| (Intercept) | 0.404 | 0.379 – 0.429 | < .001 |
| Zero-Shot | 0.059 | 0.031 – 0.088 | < .001 |
| Few-Shot | 0.089 | 0.060 – 0.119 | < .001 |
| **Random Effects** | | | |
| $\sigma^2$ | 0.07 | | |
| $\tau_{00\ \text{IdeaID}}$ | 0.01 | | |
| $\tau_{00\ \text{RespondentID}}$ | 0.02 | | |
| $\tau_{11\ \text{IdeaID.Zero-Shot}}$ | 0.04 | | |
| $\tau_{11\ \text{IdeaID.Few-Shot}}$ | 0.01 | | |
| $\tau_{11\ \text{RespondentID.Zero-Shot}}$ | 0.01 | | |
| $\tau_{11\ \text{RespondentID.Few-Shot}}$ | 0.01 | | |
| $\rho_{01}$ | -1.00 | | |
| | -0.50 | | |
| | -0.28 | | |
| | -0.06 | | |
| $N_{\text{RespondentID}}$ | 200 | | |
| $N_{\text{IdeaID}}$ | 400 | | |
| Observations | 7982 | | |
| Marginal $R^2$ / Conditional $R^2$ | 0.020 / NA | | |

**Table S3.** Estimated Difference in Purchase Intent Between GPT-4 Zero-Shot and Few-Shot

| | **Purchase Intent Rating** | | |
|---|---|---|---|
| *Predictors* | *Estimates* | *CI* | *p* |
| (Intercept) | 0.463 | 0.433 – 0.494 | < .001 |
| Few-Shot | 0.030 | -0.001 – 0.060 | 0.057 |
| **Random Effects** | | | |
| $\sigma^2$ | 0.07 | | |
| $\tau_{00\ RespondentID}$ | 0.03 | | |
| $\tau_{00\ IdeaID}$ | 0.01 | | |
| $\tau_{11\ RespondentID.Few\text{-}Shot}$ | 0.00 | | |
| $\tau_{11\ IdeaID.Few\text{-}Shot}$ | 0.02 | | |
| $\rho_{01\ RespondentID}$ | -1.00 | | |
| $\rho_{01\ IdeaID}$ | -0.64 | | |
| $N_{RespondentID}$ | 200 | | |
| $N_{IdeaID}$ | 200 | | |
| Observations | 3974 | | |
| Marginal $R^2$ / Conditional $R^2$ | 0.003 / NA | | |

**Table S4.** Estimated Difference in Purchase Intent Between Humans (Baseline) and GPT-4 Zero-Shot and Few-Shot (Akin to Table S2) Using an Alternative Weighting Scheme Following Ulrich and Eppinger (2007)

| | Purchase Intent Rating (Ulrich and Eppinger) | | |
|---|---|---|---|
| *Predictors* | *Estimates* | *CI* | *p* |
| (Intercept) | 0.076 | 0.067 – 0.085 | < .001 |
| Zero-Shot | 0.018 | 0.008 – 0.029 | 0.001 |
| Few-Shot | 0.029 | 0.018 – 0.040 | < .001 |
| **Random Effects** | | | |
| $\sigma^2$ | 0.01 | | |
| $\tau_{00\ IdeaID}$ | 0.00 | | |

| | |
|---|---|
| $\tau_{00}$ RespondentID | 0.00 |
| $\tau_{11}$ IdeaID.Zero-Shot | 0.00 |
| $\tau_{11}$ IdeaID.Few-Shot | 0.00 |
| $\tau_{11}$ RespondentID.Zero-Shot | 0.00 |
| $\tau_{11}$ RespondentID.Few-Shot | 0.00 |
| $\rho_{01}$ | -0.14 |
| | -0.74 |
| | -0.16 |
| | -0.00 |
| N RespondentID | 200 |
| N IdeaID | 400 |
| Observations | 7982 |
| Marginal $R^2$ / Conditional $R^2$ | 0.011 / NA |

**Table S5.** Estimated Difference in Purchase Intent Between Humans (baseline) and GPT-4 Zero-Shot and Few-Shot (akin to Table S2) Without Random Effects

| | Purchase Intent Rating | | |
|---|---|---|---|
| *Predictors* | *Estimates* | *CI* | *p* |
| (Intercept) | 0.404 | 0.394 – 0.414 | < .001 |
| Zero-Shot | 0.060 | 0.043 – 0.077 | < .001 |
| Few-Shot | 0.086 | 0.069 – 0.104 | < .001 |
| Observations | 7982 | | |
| $R^2$ / $R^2$ adjusted | 0.014 / 0.013 | | |

**Table S6.** Estimated Difference in Purchase Intent Between Humans (baseline) and GPT-4 Zero-Shot and Few-Shot (akin to Table S2) Controlling for Word Count

| | Purchase Intent Rating | | |
|---|---|---|---|
| *Predictors* | *Estimates* | *CI* | *p* |
| (Intercept) | 0.397 | 0.359 – 0.434 | < .001 |
| Zero-Shot | 0.059 | 0.030 – 0.088 | < .001 |
| Few-Shot | 0.088 | 0.058 – 0.118 | < .001 |
| Word Count | 0.000 | 0.000 – 0.001 | 0.617 |
| **Random Effects** | | | |
| $\sigma^2$ | 0.07 | | |
| $\tau_{00\ IdeaID}$ | 0.01 | | |
| $\tau_{00\ RespondentID}$ | 0.02 | | |
| $\tau_{11\ IdeaID.Zero\text{-}Shot}$ | 0.03 | | |
| $\tau_{11\ IdeaID.Few\text{-}Shot}$ | 0.00 | | |
| $\tau_{11\ RespondentID.Zero\text{-}Shot}$ | 0.01 | | |
| $\tau_{11\ RespondentID.Few\text{-}Shot}$ | 0.01 | | |
| $\rho_{01}$ | -0.96 | | |
| | -0.42 | | |
| | -0.28 | | |
| | -0.06 | | |
| $N_{RespondentID}$ | 200 | | |
| $N_{IdeaID}$ | 400 | | |
| Observations | 7982 | | |
| Marginal $R^2$ / Conditional $R^2$ | 0.020 / NA | | |

**Table S7.** Estimated Difference in Purchase Intent Between Original Human Ideas and Ideas Rephrased by AI

| | **Purchase Intent Rating** | | |
|---|---|---|---|
| *Predictors* | *Estimates* | *CI* | *p* |
| (Intercept) | 0.556 | 0.530 – 0.581 | < .001 |
| Condition [rephrased] | 0.009 | -0.005 – 0.023 | .205 |

| | |
|---|---|
| **Random Effects** | |
| $\sigma^2$ | 0.07 |
| $\tau_{00}$ RespondentID | 0.02 |
| $\tau_{00}$ IdeaID | 0.01 |
| $\tau_{11}$ RespondentID.Conditionrephrased | 0.00 |
| $\tau_{11}$ IdeaID.Conditionrephrased | 0.00 |
| $\rho_{01}$ RespondentID | -0.02 |
| $\rho_{01}$ IdeaID | -0.40 |
| ICC | 0.30 |
| N IdeaID | 200 |
| N RespondentID | 203 |
| Observations | 8120 |
| Marginal $R^2$ / Conditional $R^2$ | 0.000 / 0.303 |

**Table S8.** Estimated Difference in Purchase Intent Between Original Human Ideas and Ideas Rephrased by AI (akin to Table S6) Controlling for Word Count

| | Purchase Intent Rating | | |
|---|---|---|---|
| *Predictors* | *Estimates* | *CI* | *p* |
| (Intercept) | 0.557 | 0.531 – 0.582 | < .001 |
| Condition [rephrased] | 0.007 | -0.009 – 0.023 | .374 |
| Word Count | 0.0001† | -0.0002 – 0.0005† | 0.517 |
| **Random Effects** | | | |
| $\sigma^2$ | 0.07 | | |
| $\tau_{00}$ RespondentID | 0.02 | | |
| $\tau_{00}$ IdeaID | 0.01 | | |
| $\tau_{11}$ RespondentID.Conditionrephrased | 0.00 | | |

| | |
|---|---|
| $\tau_{11}$ IdeaID.Conditionrephrased | 0.00 |
| $\rho_{01}$ RespondentID | -0.02 |
| $\rho_{01}$ IdeaID | -0.40 |
| ICC | 0.30 |
| N IdeaID | 200 |
| N RespondentID | 203 |
| Observations | 8120 |
| Marginal $R^2$ / Conditional $R^2$ | 0.000 / 0.303 |

*Notes*. †We report an additional digit for extra precision.

**Table S9.** Estimated Difference in Idea Novelty Between Humans (baseline) and GPT-4 Zero-Shot and Few-Shot

| | **Novelty Rating** | | |
|---|---|---|---|
| *Predictors* | *Estimates* | *CI* | *p* |
| (Intercept) | 0.408 | 0.386 – 0.430 | < .001 |
| Zero-Shot | -0.038 | -0.066 – -0.010 | 0.008 |
| Few-Shot | -0.049 | -0.077 – -0.021 | 0.001 |
| **Random Effects** | | | |
| $\sigma^2$ | 0.05 | | |
| $\tau_{00}$ IdeaID | 0.01 | | |
| $\tau_{00}$ RespondentID | 0.01 | | |
| $\tau_{11}$ IdeaID.Zero-Shot | 0.04 | | |
| $\tau_{11}$ IdeaID.Few-Shot | 0.03 | | |
| $\tau_{11}$ RespondentID.Zero-Shot | 0.01 | | |
| $\tau_{11}$ RespondentID.Few-Shot | 0.01 | | |
| $\rho_{01}$ | -1.00 | | |
| | -1.00 | | |
| | 0.06 | | |

| | |
|---|---|
| | 0.14 |
| N $_{RespondentID}$ | 201 |
| N $_{IdeaID}$ | 400 |
| Observations | 8023 |
| Marginal $R^2$ / Conditional $R^2$ | 0.009 / NA |

**Table S10.** Estimated Difference in Idea Novelty Between GPT-4 Zero-Shot and Few-Shot

| | Novelty Rating | | |
|---|---|---|---|
| *Predictors* | *Estimates* | *CI* | *p* |
| (Intercept) | 0.370 | 0.341 – 0.399 | < .001 |
| Few-Shot | -0.011 | -0.039 – 0.017 | 0.430 |
| **Random Effects** | | | |
| $\sigma^2$ | 0.05 | | |
| $\tau_{00\ RespondentID}$ | 0.02 | | |
| $\tau_{00\ IdeaID}$ | 0.01 | | |
| $\tau_{11\ RespondentID.Few\text{-}Shot}$ | 0.00 | | |
| $\tau_{11\ IdeaID.Few\text{-}Shot}$ | 0.02 | | |
| $\rho_{01\ RespondentID}$ | -0.41 | | |
| $\rho_{01\ IdeaID}$ | -0.87 | | |
| ICC | 0.36 | | |
| N $_{RespondentID}$ | 201 | | |
| N $_{IdeaID}$ | 200 | | |
| Observations | 3955 | | |
| Marginal $R^2$ / Conditional $R^2$ | 0.000 / 0.363 | | |

**Table S11.** Quantile Regression Results for Estimated Difference in Purchase Intent Between Humans and AI (GPT-4 Zero-Shot and GPT-4 Few-Shot)

| Quantile | Intercept | Source AI (95% CI) | *P*-Value |
|---|---|---|---|
| 0.1 | 1 | 0.300 [0.128, 0.472] | < .001 |
| 0.2 | 1.231 | 0.291 [0.153, 0.429] | < .001 |
| 0.3 | 1.389 | 0.278 [0.149, 0.406] | < .001 |
| 0.4 | 1.55 | 0.262 [0.139, 0.386] | < .001 |
| 0.5 | 1.667 | 0.228 [0.109, 0.347] | < .001 |
| 0.6 | 1.79 | 0.210 [0.096, 0.325] | < .001 |
| 0.7 | 1.882 | 0.261 [0.138, 0.383] | < .001 |
| 0.8 | 1.955 | 0.445 [0.328, 0.563] | < .001 |
| 0.9 | 2.182 | 0.318 [0.191, 0.446] | < .001 |

*Notes.* The regression model considered quantiles 0.1 to 0.9 with a 0.1 step. For each quantile, it estimated MeanRating ~ SourceAI. SourceAI is a dummy variable that indicates whether the idea source was a student (SourceAI=0) or GPT-4 (SourceAI=1). A positive value for SourceAI indicates that ideas by GPT-4 performed better than human ideas. Negative values indicate the opposite.

**Table S12.** Similarity Analysis (Average Pairwise Cosine Similarity) on Student Pool and Two GPT-4 Pools (Zero-Shot and Few-Shot)

| **Group** | **Model 1** | **Model 2** | **Model 3** | **Model 4** |
|---|---|---|---|---|
| Human Ideas | 0.221***† | 0.221***† | 0.221***† | |
| GPT-4 zero-shot | 0.193*** | 0.193*** | | 0.414***† |
| GPT-4 few-shot | 0.201*** | | 0.207*** | 0.014* |
| *N* | 400 | 300 | 300 | 200 |

*Notes.* * $p < .05$, ** $p < .01$, *** $p < .001$. † Regression baseline. Model 1 includes ideas from humans and both GPT-4 conditions with fixed effects for both and uses human ideas as the baseline. Model 2 compares humans and GPT-4 zero-shot. Model 3 compares humans and GPT-4 few-shot. Model 4 compares GPT-4 zero-shot and GPT-4 few-shot and uses GPT-4 zero-shot as the baseline.

**Table S13.** Estimated Effect of Cosine Distance to Nearest LLM Idea on Human Idea Quality and Novelty (OLS, Human Ideas Only)

| | **Purchase Intent Rating** | | | **Novelty** | | |
|---|---|---|---|---|---|---|
| *Predictors* | *Estimates* | *CI* | *p* | *Estimates* | *CI* | *p* |
| (Intercept) | 0.556 | 0.483 – 0.629 | < .001 | 0.451 | 0.371 – 0.531 | < .001 |
| Distance to nearest LLM idea | -0.288 | -0.424 – -0.153 | < .001 | -0.078 | -0.226 – 0.071 | .303 |

| | | |
|---|---|---|
| Observations | 200 | 200 |
| $R^2$ / $R^2$ adjusted | 0.081 / 0.077 | 0.005 / 0.000 |

**Table S14.** Similarity Analysis (Cosine Similarity to Centroid) Across Four Different Studies Between Human and LLM-Assisted Generations

| Comparison | Human Group Result | GPT-aided Result | Effect Size | Statistics |
|---|---|---|---|---|
| | | **Present work - Study 1** | | |
| GPT-4 zero-shot | 0.466 | 0.65 | $B = 0.184$<br>95% CI [0.162, 0.206]<br>$d = 2.015$ | $t(298) = 16.46$<br>$p < .001$ |
| GPT-4 few-shot | 0.466 | 0.639 | $B = 0.173$<br>95% CI [0.151, 0.195]<br>$d = 1.901$ | $t(298) = 15.52$<br>$p < .001$ |
| | | **Boussioux (Problems)** | | |
| Single problem | 0.492 | 0.669 | $B = 0.177$<br>95% CI [0.114, 0.240]<br>$d = 1.27$ | $t(82) = 5.58$<br>$p < .001$ |
| Single problem (random persona) | 0.494 | 0.739 | $B = 0.247$<br>95% CI [0.183, 0.310]<br>$d = 1.76$ | $t(82) = 7.71$<br>$p < .001$ |
| Single problem (expert persona) | 0.492 | 0.656 | $B = 0.164$<br>95% CI [0.100, 0.228]<br>$d = 1.17$ | $t(82) = 5.12$<br>$p < .001$ |
| Multiple problems | 0.492 | 0.754 | $B = 0.262$<br>95% CI [0.199, 0.325]<br>$d = 1.879$ | $t(82) = 8.25$<br>$p < .001$ |
| Multiple problems (random persona) | 0.492 | 0.794 | $B = 0.302$<br>95% CI [0.239, 0.365]<br>$d = 2.18$ | $t(82) = 9.57$<br>$p < .001$ |
| Multiple problems (expert persona) | 0.492 | 0.744 | $B = 0.252$<br>95% CI [0.189, 0.315]<br>$d = 1.81$ | $t(82) = 7.96$<br>$p < .001$ |
| | | **Boussioux (Solutions)** | | |
| Single solution | 0.609 | 0.740 | $B = 0.131$<br>95% CI [0.092, 0.169]<br>$d = 1.54$ | $t(82) = 6.78$<br>$p < .001$ |
| Single solution (random persona) | 0.609 | 0.779 | $B = 0.170$<br>95% CI [0.132, 0.208]<br>$d = 2.02$ | $t(82) = 8.86$<br>$p < .001$ |
| Single solution (expert persona) | 0.609 | 0.738 | $B = 0.128$<br>95% CI [0.090, 0.167]<br>$d = 1.50$ | $t(82) = 6.59$<br>$p < .001$ |
| Multiple solutions | 0.609 | 0.763 | $B = 0.154$<br>95% CI [0.116, 0.192]<br>$d = 1.83$ | $t(82) = 8.02$<br>$p < .001$ |
| Multiple solutions (random persona) | 0.609 | 0.817 | $B = 0.208$<br>95% CI [0.170, 0.245]<br>$d = 2.51$ | $t(82) = 11.00$<br>$p < .001$ |
| Multiple solutions (expert persona) | 0.609 | 0.778 | $B = 0.169$<br>95% CI [0.131, 0.206]<br>$d = 2.03$ | $t(82) = 8.92$<br>$p < .001$ |
| | | **Hubert** | | |
| GPT-4 | 0.334 | 0.519 | $B = 0.185$<br>95% CI [0.179, 0.191]<br>$d = 1.930$ | $t(3489) = 57.01$<br>$p < .001$ |
| | | **Stevenson** | | |
| GPT-3 | 0.414 | 0.435 | $B = 0.021$<br>95% CI [0.010, 0.032]<br>$d = 0.191$ | $t(1564) = 3.78$<br>$p < .001$ |

*Notes.* Diversity comparisons using cosine similarity to centroid of all other embeddings (method 2). Smaller values indicate greater diversity. *P*-values are based on two-sided *t*-tests and are uncorrected.

**Table S15.** Similarity Analysis (Minimum Spanning Tree) Across Four Different Studies Between Human and LLM-Assisted Generations

| Comparison | Human Group Result | GPT-aided Result | Effect Size | Statistics |
|---|---|---|---|---|
| | | **Present work - Study 1** | | |
| GPT-4 zero-shot | 0.481 | 0.336 | Difference = 0.145<br>95% CI [0.110, 0.179]<br>$d = 1.75$ | $p < .001$ |
| GPT-4 few-shot | 0.481 | 0.311 | Difference = 0.170<br>95% CI [0.136, 0.204]<br>$d = 2.06$ | $p < .001$ |
| | | **Boussioux (Problems)** | | |
| Single problem | 0.500 | 0.352 | Difference = 0.147<br>95% CI [0.081, 0.214]<br>$d = 1.57$ | $p < .001$ |
| Single problem (random persona) | 0.500 | 0.288 | Difference = 0.211<br>95% CI [0.134, 0.289]<br>$d = 2.29$ | $p < .001$ |
| Single problem (expert persona) | 0.500 | 0.364 | Difference = 0.135<br>95% CI [0.064, 0.206]<br>$d = 1.34$ | $p = .002$ |
| Multiple problems | 0.500 | 0.256 | Difference = 0.243<br>95% CI [0.162, 0.325]<br>$d = 2.71$ | $p < .001$ |
| Multiple problems (random persona) | 0.500 | 0.228 | Difference = 0.272<br>95% CI [0.187, 0.357]<br>$d = 2.98$ | $p < .001$ |
| Multiple problems (expert persona) | 0.500 | 0.277 | Difference = 0.223<br>95% CI [0.146, 0.300]<br>$d = 2.42$ | $p < .001$ |
| | | **Boussioux (Solutions)** | | |
| Single solution | 0.415 | 0.265 | Difference = 0.149<br>95% CI [0.093, 0.207]<br>$d = 1.94$ | $p < .001$ |
| Single solution (random persona) | 0.415 | 0.248 | Difference = 0.167<br>95% CI [0.109, 0.225]<br>$d = 2.33$ | $p < .001$ |
| Single solution (expert persona) | 0.415 | 0.265 | Difference = 0.150<br>95% CI [0.093, 0.201]<br>$d = 1.97$ | $p < .001$ |
| Multiple solutions | 0.415 | 0.248 | Difference = 0.167<br>95% CI [0.109, 0.225]<br>$d = 2.35$ | $p < .001$ |
| Multiple solutions (random persona) | 0.415 | 0.200 | Difference = 0.215<br>95% CI [0.151, 0.279]<br>$d = 3.24$ | $p < .001$ |
| Multiple solutions (expert persona) | 0.415 | 0.236 | Difference = 0.179<br>95% CI [0.121, 0.238]<br>$d = 2.58$ | $p < .001$ |
| | | **Hubert** | | |
| GPT-4 | 0.413 | 0.290 | Difference = 0.123<br>95% CI [0.114, 0.132]<br>$d = 1.25$ | $p < .001$ |
| | | **Stevenson** | | |
| GPT-3 | 0.371 | 0.230 | Difference = 0.141<br>95% CI [0.122, 0.161]<br>$d = 0.792$ | $p < .001$ |

*Notes.* Diversity comparisons using minimum spanning tree (method 3). Greater values indicate greater diversity. Results are based on 1,000 simulations. *P*-values are based on two-sided *t*-tests and are uncorrected.

**Table S16.** Similarity Analysis (Average Pairwise Cosine Similarity) Across All Four Studies (Five Datasets)

| | Similarity Rating | | |
|---|---|---|---|
| *Predictors* | *Estimates* | *CI* | *p* |
| (Intercept) | 0.063 | 0.046 – 0.080 | .002 |
| AI | 0.090 | 0.038 – 0.142 | .028 |
| **Random Effects** | | | |
| $\sigma^2$ | < 0.01 | | |
| $\tau_{00\ \text{dataset}}$ | < 0.01 | | |
| $\tau_{11\ \text{dataset.AI}}$ | < 0.01 | | |
| $\rho_{01\ \text{dataset}}$ | -0.64 | | |
| ICC | 0.67 | | |
| $N_{\text{dataset}}$ | 5 | | |
| Observations | 5925 | | |
| Marginal $R^2$ / Conditional $R^2$ | 0.485 / 0.832 | | |

**Table S17.** Similarity Analysis Across Four Studies Between Human and LLM-Assisted Generations

| Comparison | Human Group Result | AI-aided Result | Effect Size | Statistics |
|---|---|---|---|---|
| **Present work - Study 1** | | | | |
| GPT-4 zero-shot | 0.221 | 0.414 | *B* = 0.193<br>95% CI [0.182, 0.204]<br>*d* = 4.151 | *t*(298) = 33.89<br>*p* < .001 |
| GPT-4 few-shot | 0.221 | 0.428 | *B* = 0.207<br>95% CI [0.196, 0.219]<br>*d* = 4.407 | *t*(298) = 35.98<br>*p* < .001 |
| **Boussioux (Problems)** | | | | |
| Single problem | 0.254 | 0.465 | *B* = 0.210<br>95% CI [0.177, 0.244]<br>*d* = 2.877 | *t*(82) = 12.64<br>*p* < .001 |
| Single problem (random persona) | 0.254 | 0.56 | *B* = 0.306<br>95% CI [0.271, 0.340]<br>*d* = 4.058 | *t*(82) = 17.82<br>*p* < .001 |
| Single problem (expert persona) | 0.254 | 0.448 | *B* = 0.194<br>95% CI [0.160, 0.228]<br>*d* = 2.607 | *t*(82) = 11.45<br>*p* < .001 |
| Multiple problems | 0.254 | 0.582 | *B* = 0.328<br>95% CI [0.294, 0.362]<br>*d* = 4.403 | *t*(82) = 19.33<br>*p* < .001 |
| Multiple problems (random persona) | 0.254 | 0.642 | *B* = 0.388<br>95% CI [0.354, 0.421]<br>*d* = 5.234 | *t*(82) = 23.00<br>*p* < .001 |
| Multiple problems (expert persona) | 0.254 | 0.568 | *B* = 0.314<br>95% CI [0.280, 0.347]<br>*d* = 4.233 | *t*(82) = 18.59<br>*p* < .001 |
| | | **Boussioux (Solutions)** | | |
| Single solution | 0.382 | 0.562 | *B* = 0.180<br>95% CI [0.155, 0.204]<br>*d* = 3.325 | *t*(82) = 14.60<br>*p* < .001 |
| Single solution (random persona) | 0.382 | 0.62 | *B* = 0.238<br>95% CI [0.213, 0.262]<br>*d* = 4.378 | *t*(82) = 19.22<br>*p* < .001 |
| Single solution (expert persona) | 0.382 | 0.559 | *B* = 0.176<br>95% CI [0.152, 0.201]<br>*d* = 3.225 | *t*(82) = 14.16<br>*p* < .001 |
| Multiple solutions | 0.382 | 0.596 | *B* = 0.214<br>95% CI [0.190, 0.239]<br>*d* = 3.950 | *t*(82) = 17.35<br>*p* < .001 |
| Multiple solutions (random persona) | 0.382 | 0.678 | *B* = 0.296<br>95% CI [0.272, 0.320]<br>*d* = 5.538 | *t*(82) = 24.32<br>*p* < .001 |
| Multiple solutions (expert persona) | 0.382 | 0.618 | *B* = 0.236<br>95% CI [0.212, 0.260]<br>*d* = 4.435 | *t*(82) = 19.48<br>*p* < .001 |
| | | **Hubert** | | |
| GPT-4 | 0.112 | 0.27 | *B* = 0.158<br>95% CI [0.155, 0.160]<br>*d* = 3.940 | *t*(3489) = 116.37<br>*p* < .001 |
| | | **Stevenson** | | |
| GPT-3 | 0.172 | 0.19 | *B* = 0.018<br>95% CI [0.013, 0.022]<br>*d* = 0.380 | *t*(1564) = 7.52<br>*p* < .001 |

*Notes.* Diversity comparisons using pairwise cosine similarity (method 1). Smaller values indicate greater diversity. *P*-values are based on two-sided *t*-tests and are uncorrected.

**Table S18.** Similarity Analysis (Average Pairwise Cosine Similarity) Across Different Models (Lower Is Better)

| Model | Coefficient | 95% CI | *p*-value |
|---|---|---|---|
| Intercept (GPT-4 zero-shot, Study 1) | 0.415 | [0.406, 0.424] | < .001 |
| Claude Sonnet 3.5 | -0.097 | [-0.110, -0.085] | < .001 |
| DeepSeek R1 | -0.166 | [-0.179, -0.154] | < .001 |
| DeepSeek V3 | -0.114 | [-0.127, -0.102] | < .001 |
| Gemini Pro 2.0 | -0.167 | [-0.179, -0.154] | < .001 |
| GPT-3.5 Turbo | -0.095 | [-0.107, -0.082] | < .001 |
| GPT-4 Turbo | -0.171 | [-0.183, -0.158] | < .001 |
| GPT-4 few-shot (Study 1) | 0.014 | [0.001, 0.026] | .032 |
| GPT-4o | -0.094 | [-0.106, -0.081] | < .001 |
| o3-mini | 0.091 | [0.078, 0.103] | < .001 |
| Random Subset | -0.178 | [-0.190, -0.166] | < .001 |
| Human (Study 1) | -0.188 | [-0.200, -0.176] | < .001 |

*Notes*. $N = 1{,}200$. $F(11, 1188) = 399.0$. OLS estimates using pairwise cosine-similarity (method 1). Baseline category for condition is GPT-4 zero-shot from Study 1; all estimates therefore represent relative mean differences. Positive coefficients represent greater similarity.

**Table S19.** Impact of Temperature on Idea Diversity

| Rank | Temperature | Diversity [95% CI] |
|---|---|---|
| 1 | 1.0 | 0.679 [0.670, 0.688] |

| 2 | 0.5 | 0.626 [0.617, 0.635] |
|---|---|---|
| 3 | 0.0 | 0.618 [0.609, 0.627] |

*Notes*. Higher diversity scores are better (lower similarity). See Table S20 for more regression analysis results.

**Table S20.** Similarity Analysis (Average Pairwise Cosine Similarity) Across Different Temperatures Using GPT-4o (Lower Is Better)

| **Model** | **Coefficient** | **95% CI** | ***p*-value** |
|---|---|---|---|
| Intercept (GPT-4o, temperature 0) | 0.382 | [0.373, 0.391] | < .001 |
| Temperature 0.5 | -0.008 | [-0.021, 0.005] | .247 |
| Temperature 1.0 | -0.061 | [-0.074, -0.048] | < .001 |

*Notes*. $N = 300$. $F(2, 297) = 51.23$. OLS estimates using pairwise cosine-similarity (method 1). Baseline category for condition is GPT-4o zero-shot with temperature set to 0; all estimates therefore represent relative mean differences. Positive coefficients represent greater similarity.

**Table S21.** Similarity Analysis (Average Pairwise Cosine Similarity) Across Different Diversification Strategies (Lower Is Better)

| **Model** | **Coefficient** | **95% CI** | ***p*-value** |
|---|---|---|---|
| Intercept (Human, Study 1) | 0.227 | [0.217, 0.236] | < .001 |
| GPT-4 zero-shot, Study 1 | 0.188 | [0.175, 0.201] | < .001 |
| GPT-4o zero-shot | 0.094 | [0.081, 0.107] | < .001 |
| GPT-4o Chain-of-Thought | 0.075 | [0.062, 0.088] | < .001 |
| Agent 1 (refine ideas 2 times) | 0.199 | [0.186, 0.212] | < .001 |
| Agent 2 (human-picked problem) | 0.035 | [0.022, 0.048] | < .001 |
| GPT-4o random personas | 0.016 | [0.003, 0.029] | < .001 |
| GPT-4o random sonnets | 0.087 | [0.075, 0.100] | < .001 |
| GPT-4o random constraints | 0.038 | [0.025, 0.051] | < .001 |
| GPT-4o independent sessions | 0.322 | [0.309, 0.335] | < .001 |

*Notes*. $N$ = 1000; $F(9, 990) = 459.6$. OLS estimates using pairwise cosine-similarity (method 1). Baseline category for condition are the human ideas from Study 1; all estimates therefore represent relative mean differences. Positive coefficients represent greater similarity.

## E-Companion References

Bai, H., J. G. Voelkel, S. Muldowney, J. C. Eichstaedt, R. Willer. 2025. LLM-generated messages can persuade humans on policy issues. *Nature Communications* 16(1), 6037.

Boussioux, L., J. N. Lane, M. Zhang, V. Jacimovic, K. R. Lakhani. 2024. The Crowdless Future? Generative AI and Creative Problem-Solving. *Organization Science* 35(5), 1589–1607.

Brucks, M., O. Toubia. 2025. Prompt architecture induces methodological artifacts in large language models. *PLOS ONE* 20(4), e0319159.

Cox, S. R., Y. Wang, A. Abdul, C. Von Der Weth, B. Y. Lim. 2021. Directed Diversity: Leveraging Language Embedding Distances for Collective Creativity in Crowd Ideation. *Proceedings of the 2021 CHI Conference on Human Factors in Computing Systems*. Yokohama, Japan, ACM, 1–35.

De Freitas, J., G. Nave, S. Puntoni. 2025. Ideation with Generative AI—in Consumer Research and Beyond. *Journal of Consumer Research* 52(1), 18–31.

Dell'Acqua, F., E. McFowland, E. Mollick, H. Lifshitz, K. C. Kellogg, S. Rajendran, L. Krayer, F. Candelon, K. R. Lakhani. 2026. Navigating the Jagged Technological Frontier: Field Experimental Evidence of the Effects of Artificial Intelligence on Knowledge Worker Productivity and Quality. *Organization Science* 37(2), 403–423.

Doshi, A. R., O. P. Hauser. 2024. Generative AI enhances individual creativity but reduces the collective diversity of novel content. *Science Advances* 10(28), eadn5290.

Girotra, K., C. Terwiesch, K. T. Ulrich. 2010. Idea Generation and the Quality of the Best Idea. *Management Science* 56(4), 591–605.

Haase, J., P. H. P. Hanel. 2023. Artificial muses: Generative artificial intelligence chatbots have risen to human-level creativity. *Journal of Creativity* 33(3), 100066.

Haase, J., P. H. P. Hanel, S. Pokutta. 2025, April 10. Has the Creativity of Large-Language Models peaked? An analysis of inter- and intra-LLM variability. arXiv.

Herbold, S., A. Hautli-Janisz, U. Heuer, Z. Kikteva, A. Trautsch. 2023. A large-scale comparison of human-written versus ChatGPT-generated essays. *Scientific Reports* 13(1), 18617.

Hubert, K. F., K. N. Awa, D. L. Zabelina. 2024. The current state of artificial intelligence generative language models is more creative than humans on divergent thinking tasks. *Scientific Reports* 14(1), 3440.

Joosten, J., V. Bilgram, A. Hahn, D. Totzek. 2024. Comparing the Ideation Quality of Humans With Generative Artificial Intelligence. *IEEE Engineering Management Review* 52(2), 153–164.

Kalvapalle, S. G., N. Phillips, J. Cornelissen. 2024. Entrepreneurial Pitching: A Critical Review and Integrative Framework. *Academy of Management Annals* 18(2), 550–599.

Koivisto, M., S. Grassini. 2023. Best humans still outperform artificial intelligence in a creative divergent thinking task. *Scientific Reports* 13(1), 13601.

Kornish, L. J., K. T. Ulrich. 2011. Opportunity Spaces in Innovation: Empirical Analysis of Large Samples of Ideas. *Management Science* 57(1), 107–128.

Lee, B. C., J. (Jae) Chung. 2024. An empirical investigation of the impact of ChatGPT on creativity. *Nature Human Behaviour* 8(10), 1906–1914.

Matz, S. C., J. D. Teeny, S. S. Vaid, H. Peters, G. M. Harari, M. Cerf. 2024. The potential of generative AI for personalized persuasion at scale. *Scientific Reports* 14(1), 4692.

McCoy, R. T., P. Smolensky, T. Linzen, J. Gao, A. Celikyilmaz. 2023. How Much Do Language Models Copy From Their Training Data? Evaluating Linguistic Novelty in Text Generation Using RAVEN. *Transactions of the Association for Computational Linguistics* 11: 652–670.

Meincke, L., G. Nave, C. Terwiesch. 2025. ChatGPT decreases idea diversity in brainstorming. *Nature Human Behaviour* 1–3.

Mihm, J., J. Schlapp. 2019. Sourcing Innovation: On Feedback in Contests. *Management Science* 65(2), 559–576.

Nguyen, T. 2024, November 5. Understanding Transformers via N-gram Statistics. arXiv.

OpenAI. 2023, March 14. GPT-4. GPT-4 is OpenAI's most advanced system, producing safer and more useful responses. Retrieved July 1, 2025, https://openai.com/index/gpt-4/.

Osborn, A. F. 1953. *Applied Imagination: Principles and Procedures of Creative Thinking*. New York, Charles Scribner's Sons.

Salvi, F., M. Horta Ribeiro, R. Gallotti, R. West. 2025. On the conversational persuasiveness of GPT-4. *Nature Human Behaviour* 9(8), 1645–1653.

Shakespeare, W. 1997. Shakespeare's Sonnets. Project Gutenberg eBook No. 1041. Last updated March 10, 2024. Retrieved July 1, 2026, https://www.gutenberg.org/ebooks/1041.

Shibayama, S., D. Yin, K. Matsumoto. 2021. Measuring novelty in science with word embedding. *PLOS ONE* 16(7), e0254034.

Siangliulue, P., J. Chan, S. P. Dow, K. Z. Gajos. 2016. IdeaHound: improving large-scale collaborative ideation with crowd-powered real-time semantic modeling. *Proceedings of the 29th Annual Symposium on User Interface Software and Technology*. 609–624.

Stevenson, C., I. Smal, M. Baas, R. Grasman, H. van der Maas. 2022, June 10. Putting GPT-3's Creativity to the (Alternative Uses) Test. arXiv.org. Retrieved June 14, 2025, https://arxiv.org/abs/2206.08932v1.

Terwiesch, C., K. T. Ulrich. 2009. *Innovation Tournaments: Creating and Selecting Exceptional Opportunities*. Boston, MA, Harvard Business Press.

Ulrich, K., S. Eppinger. 2007. *Product Design and Development*. McGraw-Hill Education.

Wei, J., X. Wang, D. Schuurmans, M. Bosma, B. Ichter, F. Xia, E. Chi, Q. Le, D. Zhou. 2023, January 10. Chain-of-Thought Prompting Elicits Reasoning in Large Language Models. arXiv.

Wooten, J. O., K. T. Ulrich. 2017. Idea Generation and the Role of Feedback: Evidence from Field Experiments with Innovation Tournaments. *Production and Operations Management* 26(1), 80–99.

Zhou, D., N. Schärli, L. Hou, J. Wei, N. Scales, X. Wang, D. Schuurmans, C. Cui, O. Bousquet, Q. Le, E. Chi. 2023, April 16. Least-to-Most Prompting Enables Complex Reasoning in Large Language Models. arXiv.